\pdfoutput=1
\documentclass[11pt]{article}

\usepackage[]{acl}

\usepackage{times}
\usepackage{latexsym}

\usepackage[T1]{fontenc}
\usepackage[utf8]{inputenc}

\usepackage{microtype}

\usepackage{inconsolata}

\usepackage{framed}
\usepackage{mathtools}
\usepackage{graphics}
\usepackage{graphicx}
\usepackage{amsmath}
\usepackage{amsfonts,amssymb}
\usepackage{amssymb}
\usepackage{bbm}
\usepackage[ruled, linesnumbered]{algorithm2e}
\usepackage{amsmath}
\usepackage{booktabs}
\usepackage{multirow}
\usepackage{color}
\usepackage{xcolor}
\usepackage{pifont}
\usepackage{pgfplots}
\usepackage{makecell}
\usepackage{caption}
\usepackage{subcaption}

\usepackage{threeparttable}
\usepackage{adjustbox}
\usepackage{multicol}

\usepackage{paralist}
\usepackage{array}

\usepackage{booktabs}
\usepackage[online,referable]{threeparttablex}
\usepackage{multicol}

\newcommand{\Fref}[1]{Figure~\ref{#1}}

\usepackage{arydshln}
\usepackage{float}

\title{Reinforcement Learning Enhanced LLMs: A Survey}

\author{Shuhe Wang$^{\spadesuit}$, Shengyu Zhang$^{\clubsuit}$,  Jie Zhang$^\bigstar$, Runyi Hu$^\blacktriangle$, Xiaoya Li$^{\blacklozenge}$\\ 
{\bf Tianwei Zhang$^\blacktriangle$,
 Jiwei Li$^{\clubsuit}$, Fei Wu$^{\clubsuit}$, Guoyin Wang, Eduard Hovy$^{\spadesuit}$}}

\begin{document}
\maketitle
\begin{abstract}

Reinforcement learning (RL) enhanced large language models (LLMs), particularly exemplified by DeepSeek-R1, have exhibited outstanding performance.
Depsite the effectiveness in improving LLM capabilities, its implementation remains highly complex, requiring complex algorithms, reward modeling strategies, and optimization techniques. This complexity poses challenges for researchers and practitioners in developing a systematic understanding of RL-enhanced LLMs.
Moreover, the absence of a comprehensive survey summarizing existing research on RL-enhanced LLMs has limited progress in this domain, hindering further advancements.

% This paper surveys research in the rapidly growing field of enhancing large language models (LLMs) with reinforcement learning (RL), a technique that enables LLMs to improve their performance by receiving feedback in the form of rewards based on the quality of their outputs, allowing them to generate more accurate, coherent, and contextually appropriate responses.
In this work, we are going to make a systematic review of the most up-to-date state of knowledge on RL-enhanced LLMs, attempting to consolidate and analyze the rapidly growing research in this field, helping researchers understand the current challenges and advancements.
Specifically, we (1) detail the basics of RL; (2) introduce popular RL-enhanced LLMs; (3) review researches on two widely-used reward model-based RL techniques: Reinforcement Learning from Human Feedback (RLHF) and Reinforcement Learning from AI Feedback (RLAIF); and (4) explore Direct Preference Optimization (DPO), a set of methods that bypass the reward model to directly use human preference data for aligning LLM outputs with human expectations.
We will also point out current challenges and deficiencies of existing methods and suggest some avenues for further improvements. 
Project page of this work can be found at \href{https://github.com/ShuheWang1998/Reinforcement-Learning-Enhanced-LLMs-A-Survey}{our latest repo}.

\let\thefootnote\relax\footnotetext{$^{\spadesuit}$The University of Melbourne, $^{\clubsuit}$Zhejiang University, $^{\bigstar}$CFAR and IHPC, A*STAR, Singapore, $^\blacktriangle$Nanyang Technological University, $^{\blacklozenge}$University of Washington}
\let\thefootnote\relax\footnotetext{Email: shuhewang@student.unimelb.edu.au}
\let\thefootnote\relax\footnotetext{Project page of this work can be found at: \url{https://github.com/ShuheWang1998/Reinforcement-Learning-Enhanced-LLMs-A-Survey}} 
\let\thefootnote\relax\footnotetext{\bf * The latest update was on Feb. 24, 2025 (Version 2).}
\end{abstract}

\section{Introduction}

Recently, represented by DeepSeek-R1 \cite{deepseekai2025deepseekr1incentivizingreasoningcapability}, reinforcement learning (RL) enhanced large language models (LLMs) \cite{openai2023gpt4,openai2024gpt4o,team2024gemma2,glm2024chatglm,adler2024nemotron,yang2024qwen2,ai2024yilightningtechnicalreport,kimiteam2025kimik15scalingreinforcement}, have demonstrated remarkable performance  \citep{wang2023gpt, wan2023gpt, sun2023text, sun2023pushing, giray2023prompt, zhang2023meta, long2023large, sun2023reinforcement, gao2023retrieval, paranjape2023art, sun2023query, diao2023active, wang2023sim, zhang2023multimodal, sun2023sentiment, liu2024large, yao2024tree, liu2024best, lee2024can, kambhampati2024can, wang2024rethinking}, attracting widespread attention from the research community. A key factor contributing to this success is the integration of reinforcement learning techniques, which have proven to be a powerful approach to enhancing LLM capabilities.

The process of training LLMs using RL can be divided into three main steps: (1) Reward Model Training: before fine-tuning, a reward model (or reward function) is trained to approximate human preferences and evaluate different LLM outputs; (2) Preference-Based Fine-Tuning: during each fine-tuning iteration, the LLM generates multiple responses for a given instruction, each of which is scored using the trained reward model; and (3) Policy Optimization: using RL optimization techniques, the model's weights are updated based on preference scores, improving response generation.
Incorporating reinforcement learning into LLMs allows models to adjust dynamically based on varied preference scores, rather than being confined to a single predetermined answer. This enables them to produce well-structured, contextually appropriate responses. Furthermore, RL enables direct training on human preferences, enhancing the LLM’s ability to generate creative, high-quality outputs that align more closely with human expectations.

Although reinforcement learning (RL) is highly effective in enhancing LLM capabilities, its implementation remains highly complex, requiring sophisticated algorithms, reward modeling strategies, and optimization techniques. This complexity poses challenges for researchers and practitioners in developing a systematic understanding of RL-enhanced LLMs.
Moreover, the absence of a comprehensive survey summarizing existing research on RL-enhanced LLMs has limited progress in this domain, hindering further advancements and making it difficult to consolidate knowledge and drive innovation in the field.

To address this issue, in this paper, we provide a systematic review of both the challenges and opportunities involved in reinforcement learning-enhanced LLMs. Specifically, we aim to identify the key obstacles in applying reinforcement learning to LLMs and offer a comprehensive analysis of recent efforts to overcome them:

\noindent \textbf{(1) At the reward model training level}, we analyze the evolution of reward models, including Reinforcement Learning from Human Feedback (RLHF) and Reinforcement Learning from AI Feedback (RLAIF). Our discussion covers their effectiveness, limitations, evaluation methods, and challenges, such as bias in human annotations, generation of out-of-distribution content, and issues related to human interpretability.

\noindent \textbf{(2) At preference fine-tuning level},
we explore approaches for aligning LLMs with human preferences, including direct preference optimization (DPO) and its variations. Additionally, we examine the impact of preference data collection, different optimization functions, and strategies designed to maintain safety throughout the alignment process.

By presenting a comprehensive overview of RL-enhanced LLMs, this survey aims to bridge the knowledge gap in the field, attempting to consolidate and analyze the rapidly growing research in this field, helping researchers understand the current landscape, challenges, and advancements. The rest of this survey is organized as follows:

\begin{itemize}
	\item Section \ref{sec:basics_reinforcement_learning_for_llms} presents the basics of reinforcement learning (RL) along with key terminologies, and outlines how the RL pipeline is adapted for LLMs.
	\item Section \ref{sec:popular_llms_enhanced_by_rl} introduces popular and powerful LLMs enhanced by reinforcement learning.
	\item Section \ref{sec:rlhf_reinforcement_learning_from_human_feedback} outlines the process of reinforcement learning from human feedback (RLHF), a training method that integrates reinforcement learning with human feedback to align LLMs with human values, preferences, and expectations. 
	\item Section \ref{sec:rlaif_reinforcement_learning_from_ai_feedback} reviews research on reinforcement learning from AI feedback (RLAIF), which presents a promising alternative or complement to RLHF by utilizing AI systems to provide feedback on the outputs of the LLM being trained, offering advantages in scalability, consistency, and cost-effectiveness.
	\item Section \ref{sec:analysis_of_rlhf_rlaif} provides an analysis of the challenges associated with RLHF and RLAIF.
	\item Section \ref{sec:direct_preference_optimization_dpo} discusses research on direct preference optimization (DPO), a series of methods that bypasses the reward model and directly utilizes human preference data to align LLM outputs with human expectations.
	\item Section \ref{sec:analyasis_of_dpo} summarizes the current challenges and discusses opportunities for further improvement..
\end{itemize}

\section{Basics: Reinforcement Learning for LLMs}
\label{sec:basics_reinforcement_learning_for_llms}

In this section, we first detail the basics of reinforcement learning (RL) along with key terminologies, and then outline how the RL pipeline is adapted for LLMs.

\begin{figure*}[htb]
\centering
    \includegraphics[scale=0.48]{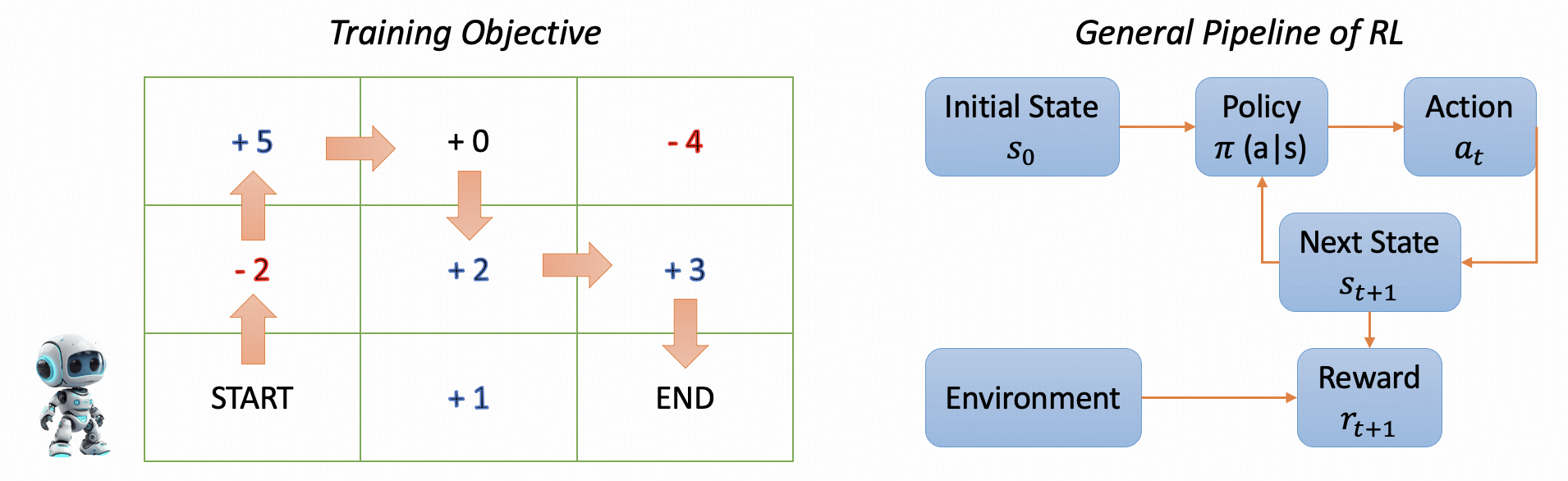}
  \caption{An example of the full process of RL. \textbf{Training Objective:} The goal is to train a robot to navigate from the bottom-left corner of a square to the top-right corner. Each grid cell is assigned a reward score, and the objective is to maximize the robot’s overall score. \textbf{General Pipeline of RL:} The agent begins in an initial state $s_0$, and at each time step $t$, it selects an action $a_{t}$ based on its current state $s_{t}$. In response, the environment transitions to a new state $s_{t+1}$, and the agent receives a reward $r_{t}$.}
  \label{fig:general_pipeline_rl}
\end{figure*}

\subsection{Basics of Reinforcement Learning}

Reinforcement Learning (RL) is a key approach in machine learning, focusing on how an agent engages with its environment to maximize cumulative rewards. Unlike supervised learning, which depends on labeled data, and unsupervised learning, which uncovers patterns in unlabeled data, RL emphasizes learning through direct feedback via trial and error. Below, we sequentially describe basic definitions and general pipeline of RL.

\subsubsection{Basic Definitions}
Here, we use the training example in Figure \ref{fig:general_pipeline_rl} to illustrate the full process of RL. In this example, our goal is to train a robot to move from the bottom-left corner of a square to the top-right corner. Additionally, each grid cell has a reward score, and we aim to maximize the robot's total score. Before delving into the training process, we first introduce some relevant terms:

\begin{itemize}
	\item \textbf{Agent:} An agent is the entity we train to make correct decisions. In this example, our goal is to train the robot to make movement decisions, so the robot is the agent.
	\item \textbf{Environment:} The environment is the external system that the agent interacts with. For our example, as the trained robot (agent) moves within the grid, the grid serves as the environment.
	\item \textbf{State:} The state represents the agent's position at each time $t$. For instance, at the beginning, at time $t_0$, the robot (agent) starts at the bottom-left corner, so the state at time $t_0$ is the bottom-left corner, represented by the coordinates $(0, 0)$.
	\item \textbf{Action(s):} Actions represent the possible choices available to the agent within the environment at each time $t$. For example, at the start, at time $t_0$, the robot (agent) can choose to move right or up, making these two actions available to the agent at $t_0$.
	\item \textbf{Reward(s):} Rewards are the signals or feedback provided by the environment to the agent based on the action it takes at each time $t$. For instance, at time $t_0$, the robot (agent) would receive a reward of +5 points for moving right, or a penalty of -1 point for moving up.
	\item \textbf{Policy:} A policy is a set of decision-making strategies that helps the agent choose an action at each time $t$. In practice, at time $t_0$, the policy represents a probability distribution that directs the robot (agent) to move right or up in order to maximize its cumulative rewards.
\end{itemize}

%\textbf{1. Agent:} An agent is the entity we train to make correct decisions. In this example, our goal is to train the robot to make movement decisions, so the robot is the agent.
%
%\textbf{2. Environment:} The environment is the external system that the agent interacts with. For our example, as the trained robot (agent) moves within the grid, the grid serves as the environment.
%
%\textbf{3. State:} The state represents the agent's position at each time $t$. For instance, at the beginning, at time $t_0$, the robot (agent) starts at the bottom-left corner, so the state at time $t_0$ is the bottom-left corner, represented by the coordinates $(0, 0)$.
%
%\textbf{4. Action(s):} Actions represent the possible choices available to the agent within the environment at each time $t$. For example, at the start, at time $t_0$, the robot (agent) can choose to move right or up, making these two actions available to the agent at $t_0$.
%
%\textbf{5. Reward(s):} Rewards are the signals or feedback provided by the environment to the agent based on the action it takes at each time $t$. For instance, at time $t_0$, the robot (agent) would receive a reward of +5 points for moving right, or a penalty of -1 point for moving up.
%
%\textbf{6. Policy:} A policy is a set of decision-making strategies that helps the agent choose an action at each time $t$. In practice, at time $t_0$, the policy represents a probability distribution that directs the robot (agent) to move right or up in order to maximize its cumulative rewards.

\subsubsection{General Pipeline of RL}

We have defined key terminologies used in RL, and in this section, we will continue to detail the general pipeline of RL. 

As illustrated in Figure \ref{fig:general_pipeline_rl}, the general reinforcement learning (RL) pipeline can be represented as a Markov Decision Process (MDP). Formally, the agent begins in an initial state $s_0$, and at each time step $t$, it selects an action $a_{t}$ based on its current state $s_{t}$. In response, the environment transitions to a new state $s_{t+1}$, and the agent receives a reward $r_{t}$. This cycle continues, with the agent’s objective being to maximize the cumulative rewards it accumulates over time.

Mapping into the specific example in Figure \ref{fig:general_pipeline_rl}, at the initial time $t_0$, the robot starts at the bottom-left corner, denoted by the position (state) $s_0$. As time progresses, at each time step $t$, the robot chooses an action $a_t$ (either moving up or moving right). This action causes the robot to transition from its current position $s_t$ to a new position $s_{t+1}$, while earning a reward $t_t$. This cycle of movement and reward collection continues until the robot reaches the desired position (state) at the top-right corner, achieving the goal of maximum cumulative rewards.

\begin{figure*}[htb]
\centering
    \includegraphics[scale=0.21]{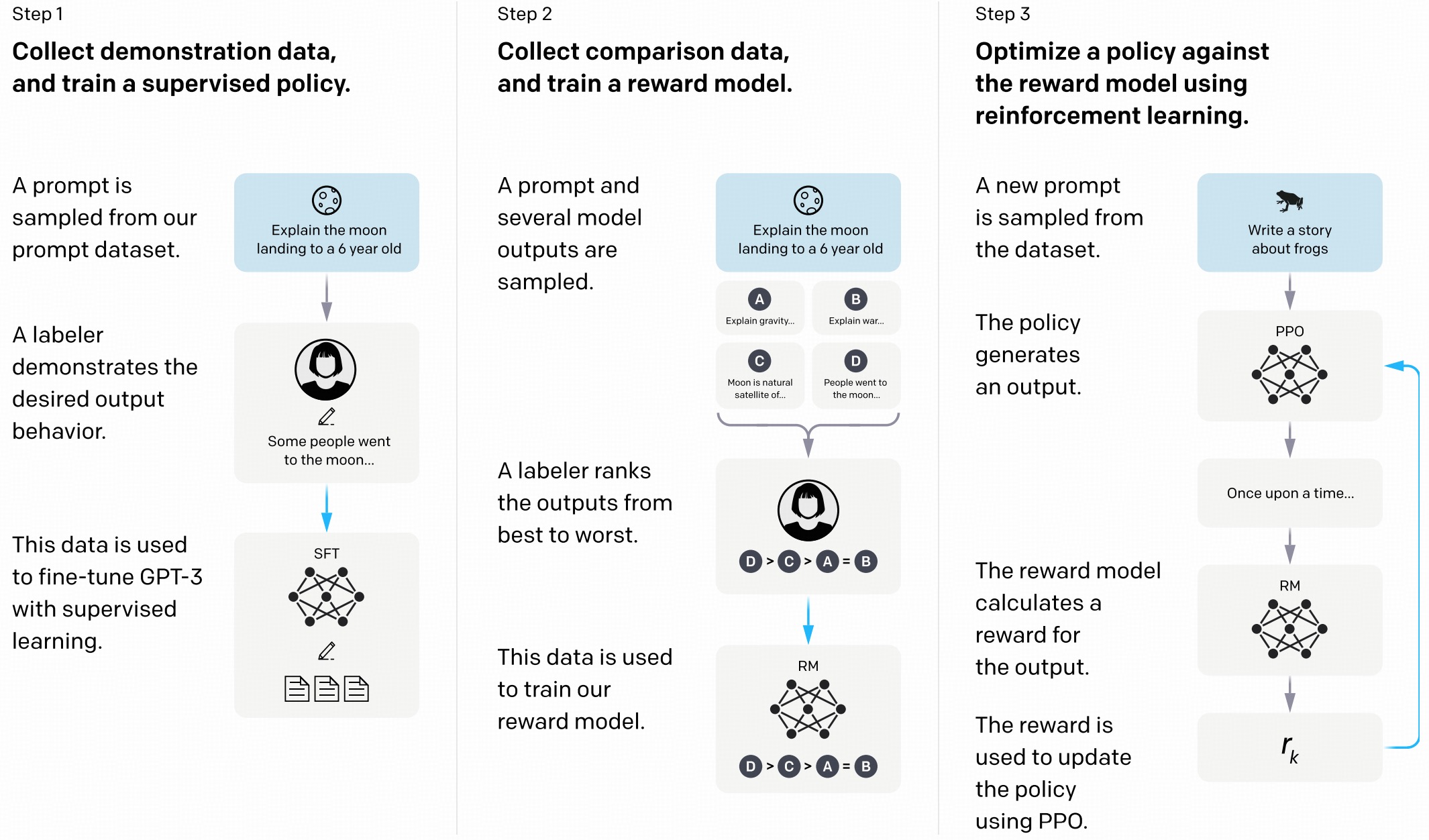}
  \caption{The framework of RL for LLMs proposed by \newcite{ouyang2022training}.}
  \label{fig:framework_rl_llms}
\end{figure*}

\subsection{RL for LLMs}

We have outlined the general framework of RL above; now we will delve into the process of fine-tuning LLMs using RL. This approach aims to align LLMs with desired behaviors, enhance their performance, and ensure that their outputs are both effective and dependable.

In reinforcement learning (RL), there are six key components:agent, environment, state, action, reward, and policy. To apply RL for fine-tuning large language models (LLMs), the first step is to map these components to the LLM framework. LLMs are highly proficient at next-token prediction, where they take a sequence of tokens as input and predict the next token based on the given context. From an RL perspective, we can view the LLM itself as the policy. The current textual sequence represents the state, and based on this state, the LLM generates an action—the next token. This action updates the state, creating a new state that incorporates the newly added token. After generating a complete textual sequence, a reward is determined by assessing the quality of the LLM's output using a pre-trained reward model.

Figure \ref{fig:framework_rl_llms} illustrates the specific RL framework for LLMs as proposed by \cite{ouyang2022training}.
\citet{ouyang2022training} starts with an instruction-tuned model trained through supervised learning, enabling it to generate structured responses to human instructions. Then, \citet{ouyang2022training} applies the following two steps:

\textbf{Step 1: Collect comparison data, and train a reward model.} \citet{ouyang2022training} collects a dataset of comparisons between outputs of the instruction-tuned model, where labelers indicate which output they prefer for a given input. Then, the collected dataset is used to train a reward model (RM) to predict the human-preferred output.

\textbf{Step 2: Optimize a policy against the reward model using PPO.} \citet{ouyang2022training} leverages the output of the RM as a scalar reward, and fine-tunes the instruction-tuned model to optimize this reward using the PPO algorithm \cite{schulman2017proximal}.

\begin{table*}[htb!]
\centering
\small
\begin{adjustbox}{max width=1.0\textwidth}
\begin{threeparttable}
\begin{tabular}{l>{\centering\arraybackslash}m{3cm}cc}
\toprule
\textbf{RL Enhanced LLMs} & \textbf{Organization} & \textbf{\# Params} & \textbf{RL Methods} \\
\midrule
DeepSeek-R1 \citep{deepseekai2025deepseekr1incentivizingreasoningcapability} & \includegraphics[height=0.5cm]{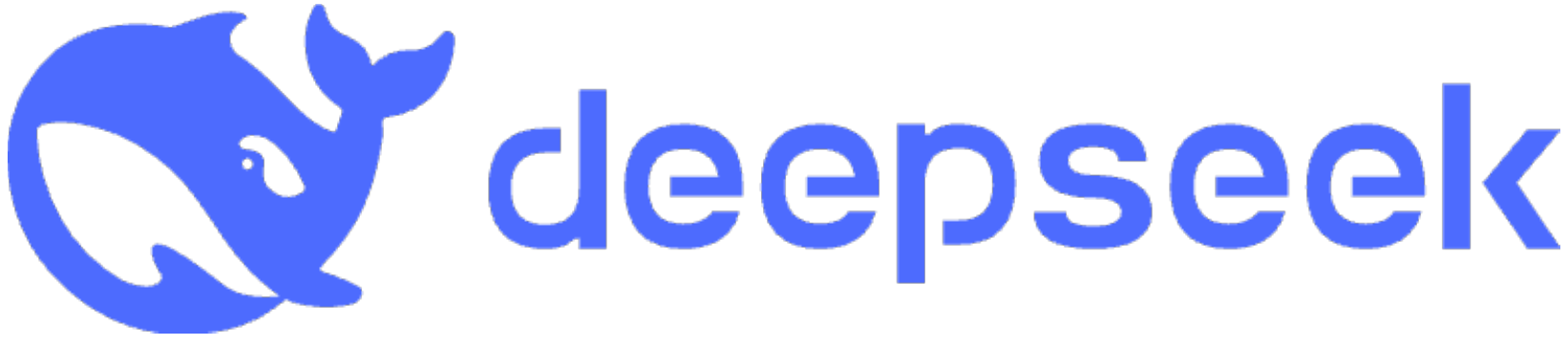} & 671B-A31B & RL through CoT \\
Kimi-k1.5 \citep{kimiteam2025kimik15scalingreinforcement} & \includegraphics[height=0.5cm]{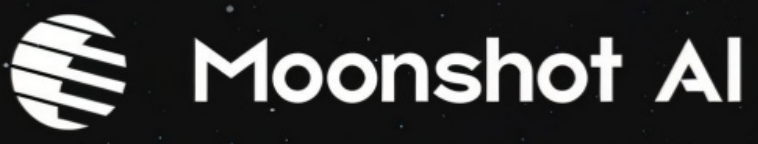} & - & RL through CoT \\
o1 \citep{o1_2024} & \includegraphics[height=0.5cm]{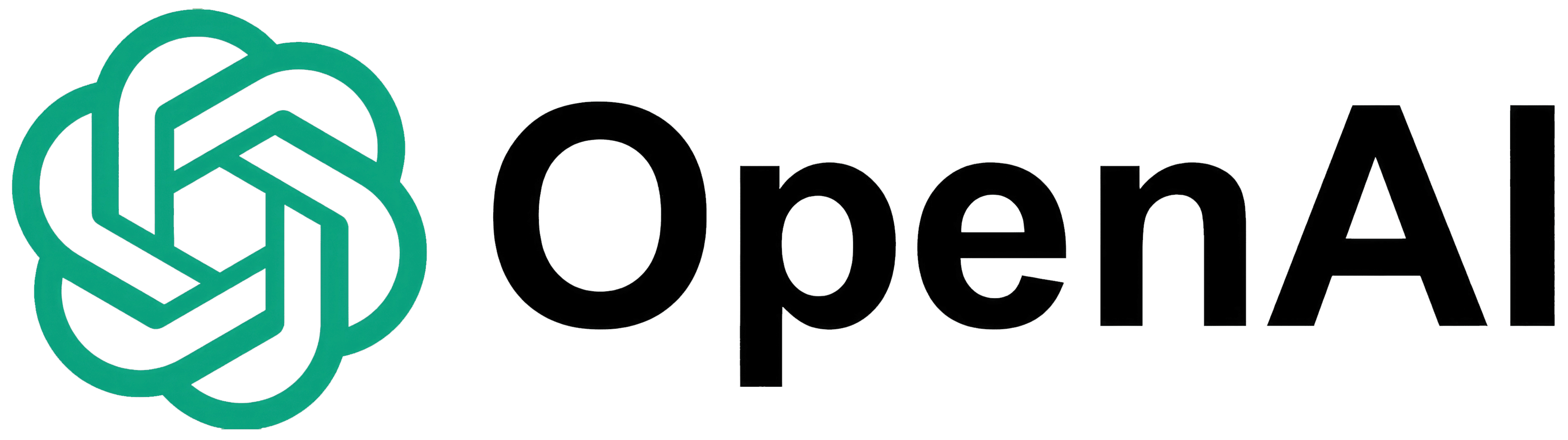} & - & RL through CoT \\
Hermes 3   \citep{teknium2024hermes} & \includegraphics[height=0.8cm]{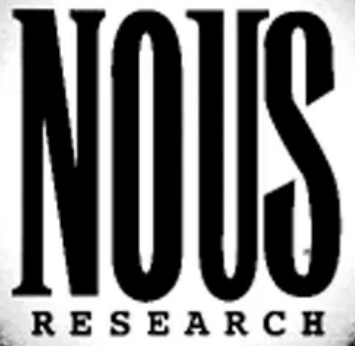} & 8B, 70B, 405B & DPO \\
Athene-70B \citep{nexusflow_athene} & \includegraphics[height=0.6cm]{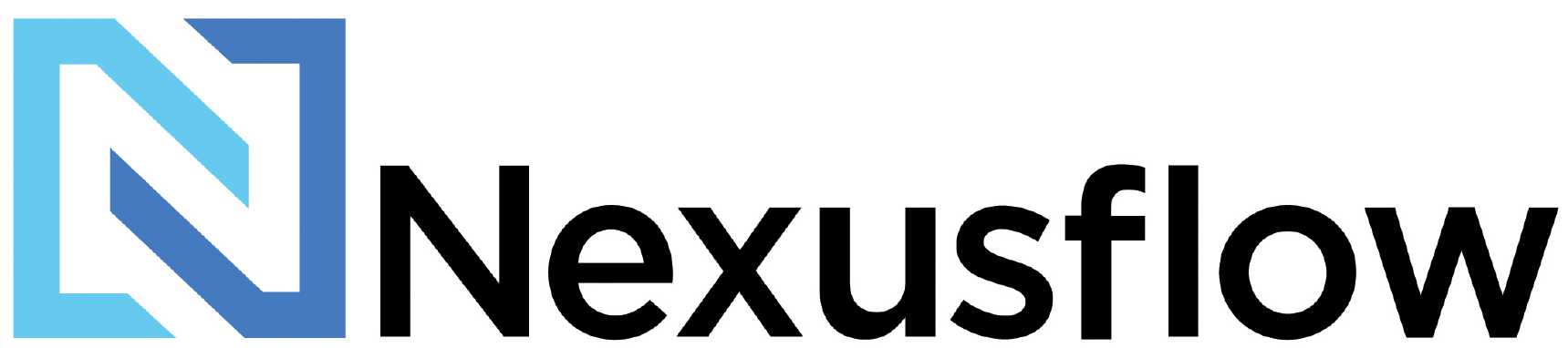} & 70B & RLHF \\
Starling-7B   \citep{zhu2024starling} & \includegraphics[height=0.6cm]{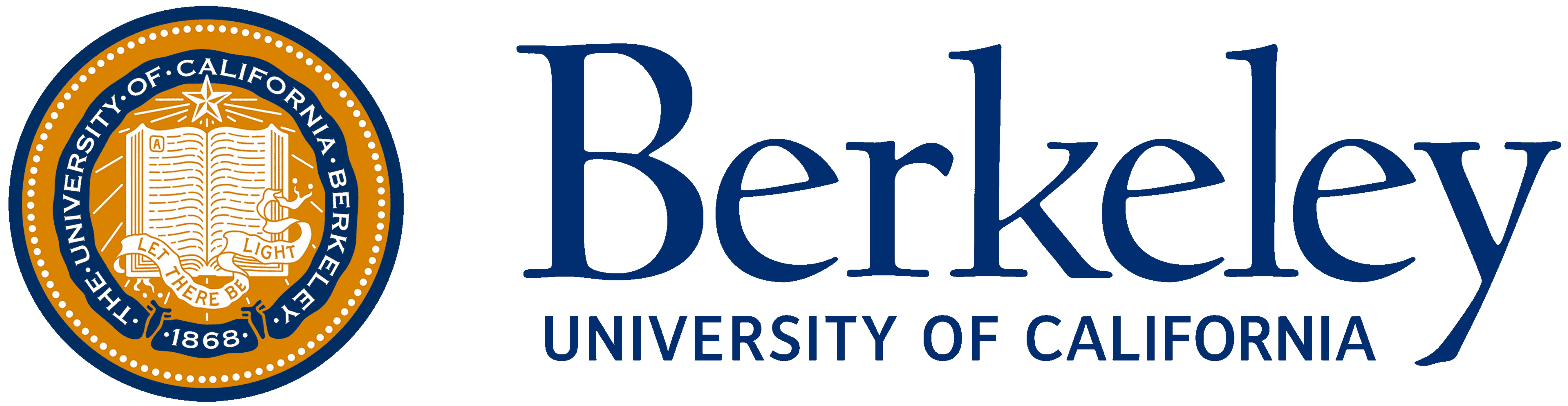} & 7B & RLAIF, PPO \\
Gemma2 \citep{team2024gemma2} & \includegraphics[height=0.5cm]{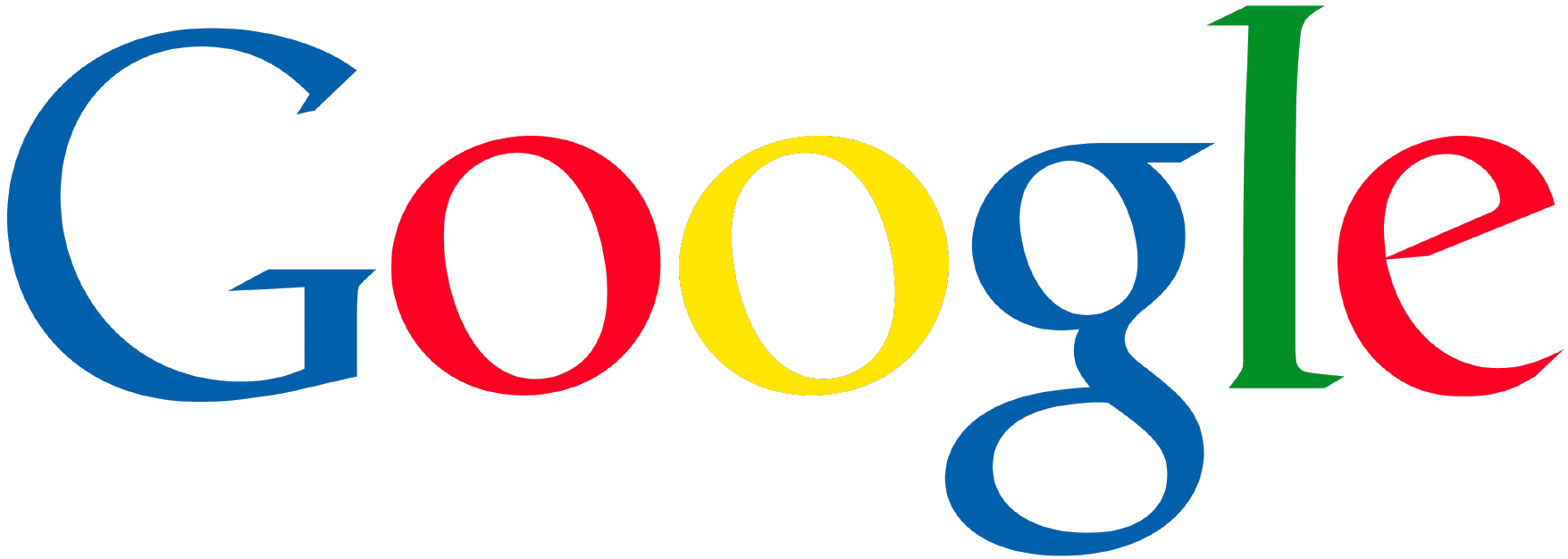} & 2B, 9B, 27B & RLHF \\
Qwen2 \citep{yang2024qwen2} & \includegraphics[height=0.38cm]{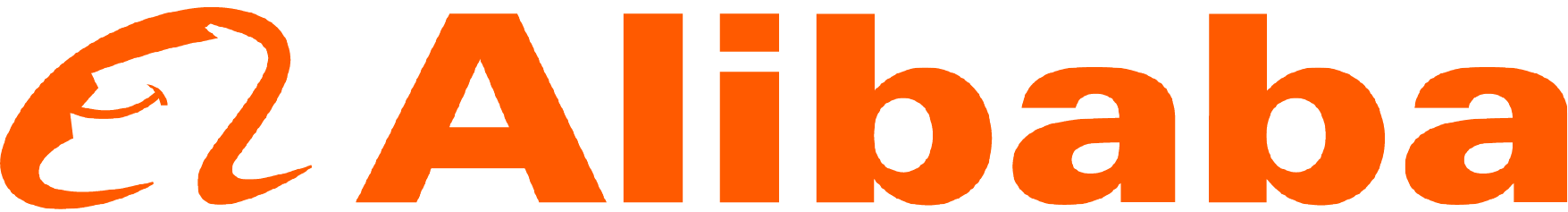} & (0.5-72)B, 57B-A14B & DPO \\
Llama 3 \citep{dubey2024llama} & \includegraphics[height=0.45cm]{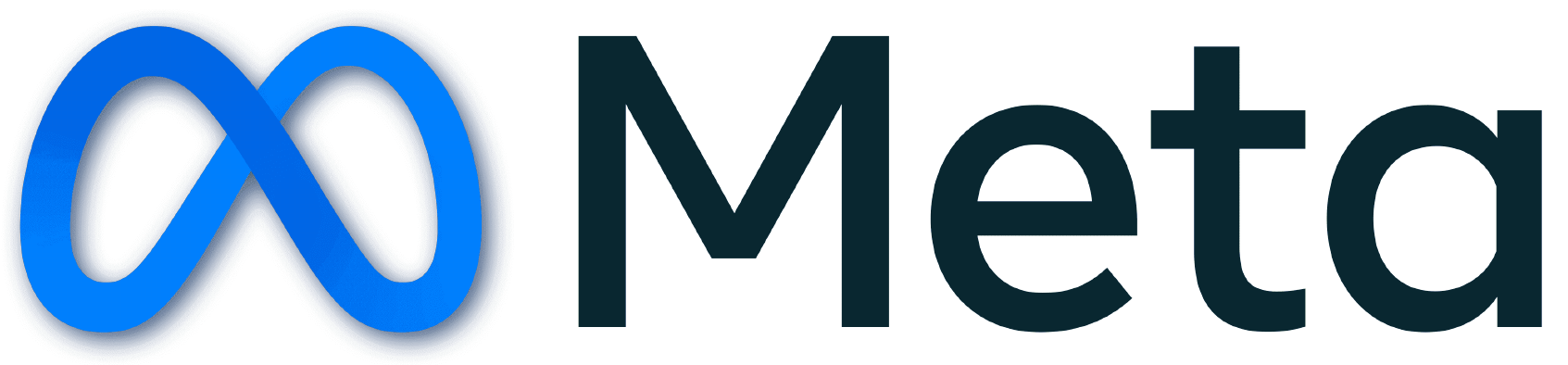} & 8B, 70B, 405B & DPO \\
Nemotron-4 340B   \citep{adler2024nemotron} & \includegraphics[height=0.45cm]{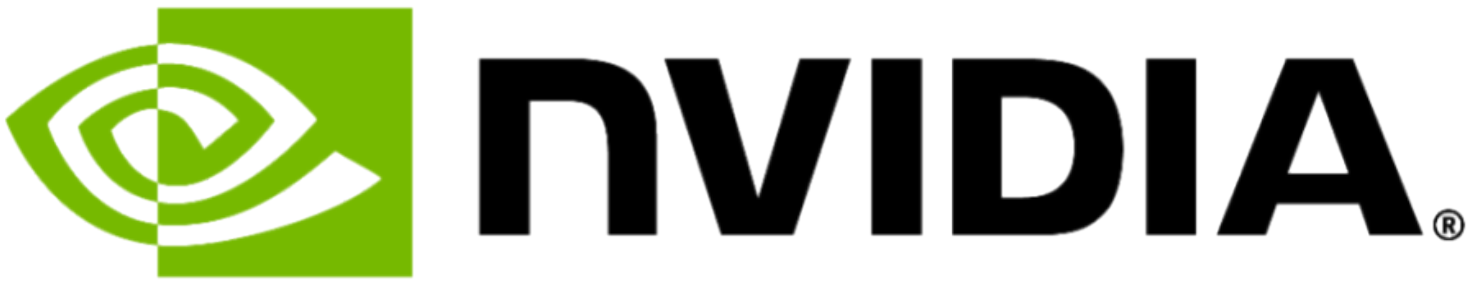} & 340B & DPO, RPO \\
ChatGLM \citep{glm2024chatglm} & \includegraphics[height=0.65cm]{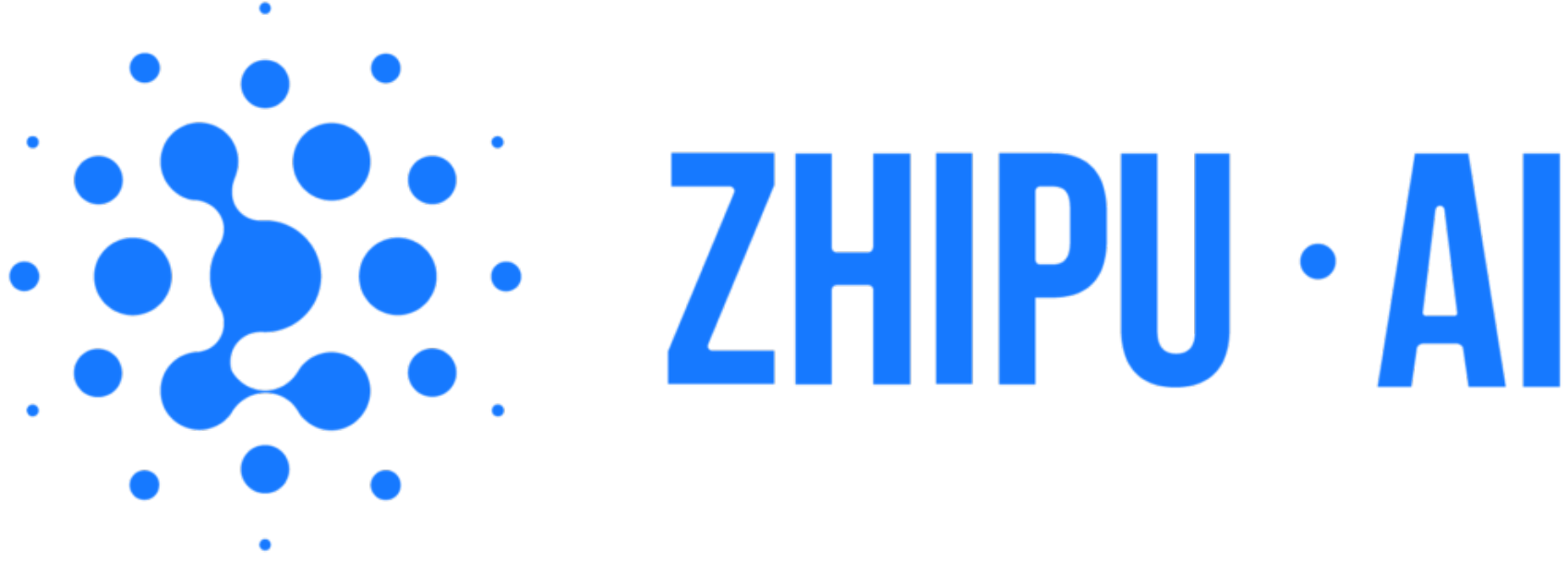}  & 6B, 9B & ChatGLM-RLHF \\
DeepSeek-V2   \citep{liu2024deepseek} & \includegraphics[height=0.55cm]{figs/sec3/deepseek.pdf}  & 236B-A21B & GRPO \\
Phi-3 \citep{abdin2024phi} & \includegraphics[height=0.5cm]{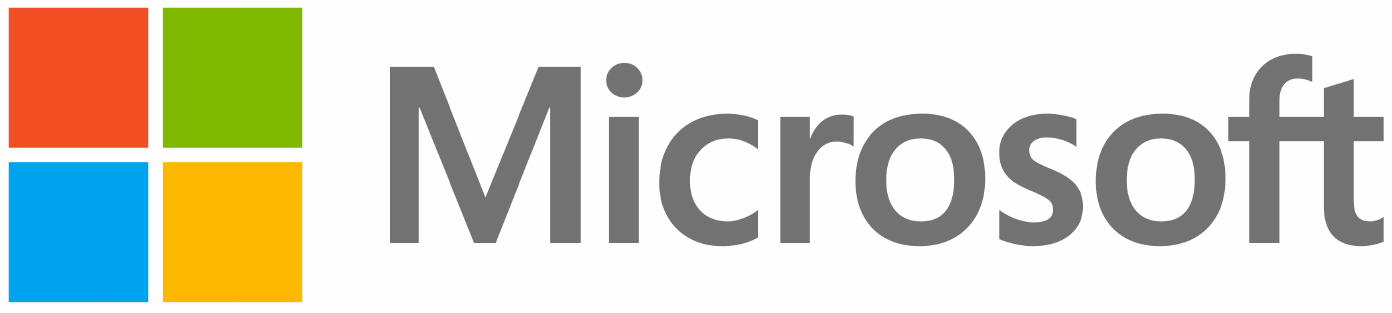}  & 3.8B, 7B, 14B & DPO \\
Zephyr   \citep{zephyr_orpo_2024} & \includegraphics[height=0.4cm]{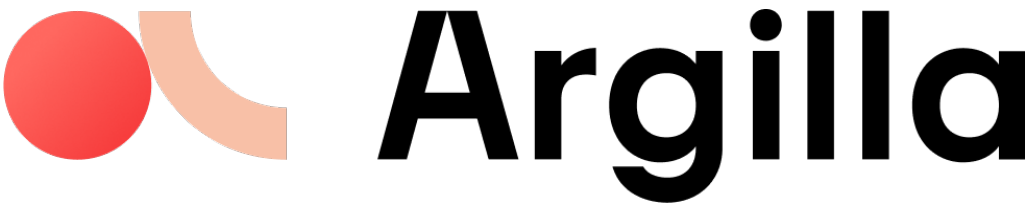} & 141B-A39B & ORPO \\
Reka \citep{team2024reka} & \includegraphics[height=0.5cm]{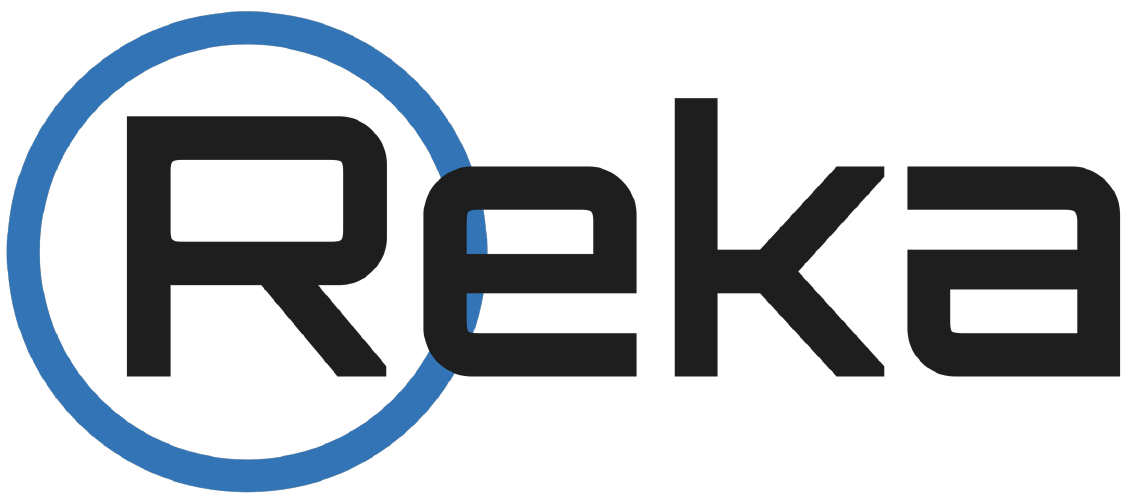}  & 7B, 21B & RLHF, PPO \\
Claude 3   \citep{anthropic_claude3_2024} & \includegraphics[height=0.35cm]{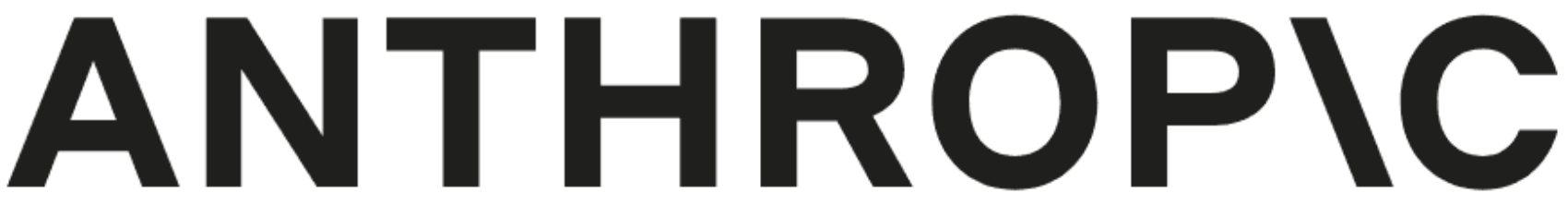}  & - & RLAIF \\
InternLM2   \citep{cai2024internlm2} & \includegraphics[height=0.95cm]{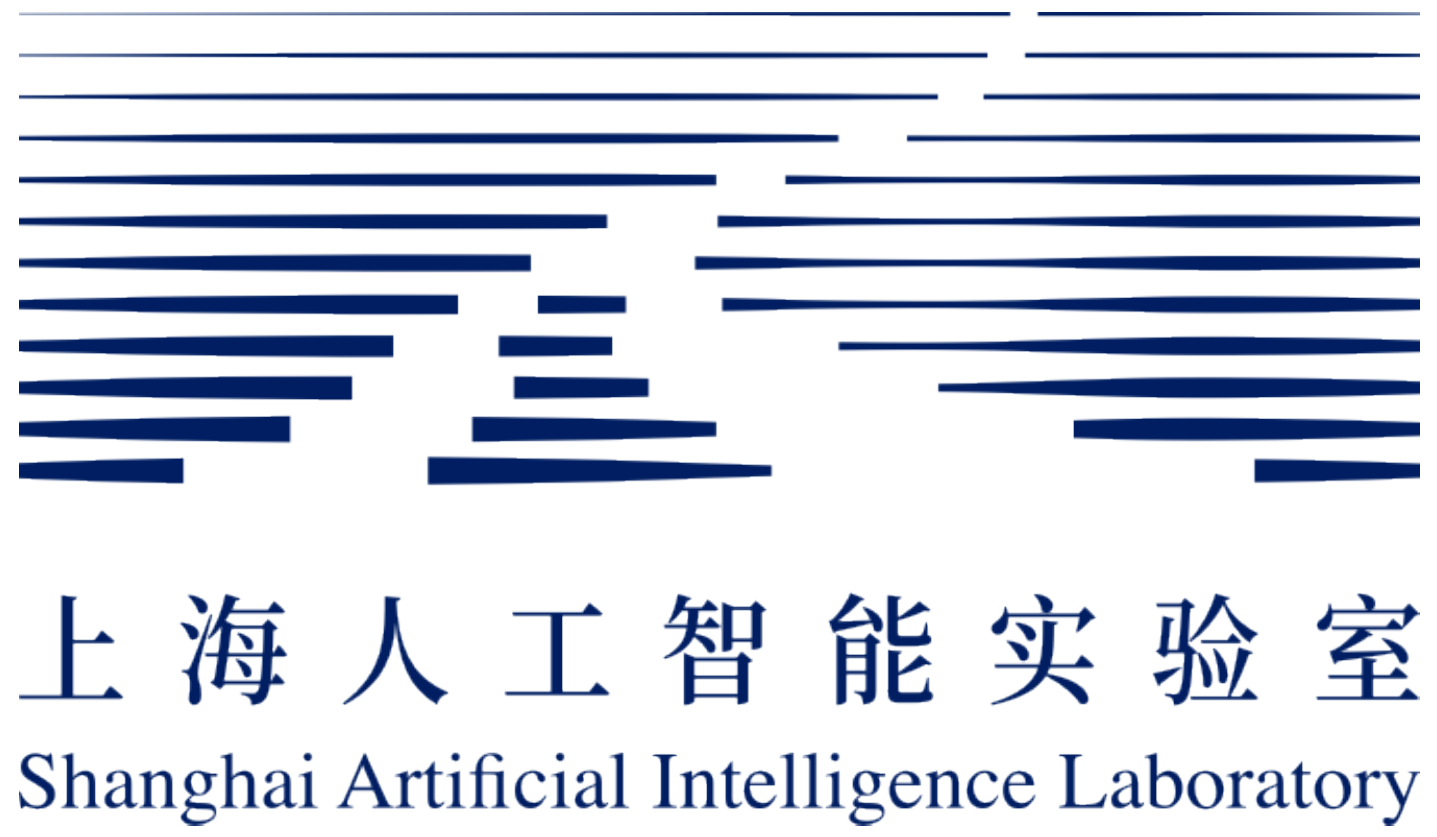} & 1.8B, 7B, 20B & RLHF, PPO \\
Gemini \citep{team2023gemini} & \includegraphics[height=0.5cm]{figs/sec3/google.pdf} & - & RLHF \\
GPT-4 \citep{openai2023gpt4} & \includegraphics[height=0.5cm]{figs/sec3/openai.pdf} & - & RLHF, PPO, RBRM \\
Instruct-GPT   \citep{ouyang2022training} & \includegraphics[height=0.5cm]{figs/sec3/openai.pdf}  & 1.3B, 6B, 175B & RLHF, PPO \\

% Yi-Lightning \citep{ai2024yilightningtechnicalreport} & \includegraphics[height=0.4cm]{figs/sec3/01ai.pdf} & - & RLHF \\
\bottomrule
\end{tabular}
\end{threeparttable}
\end{adjustbox}
% \begin{multicols}{2}
% \begin{tablenotes}
% \item[1] \label{id:1} {$^1$ https://huggingface.co/internlm/internlm2-7b} 
% \item[2] \label{id:2} {$^2$ https://huggingface.co/deepseek-ai/DeepSeek-V2} 
% \item[3] \label{id:3} {$^3$ https://huggingface.co/berkeley-nest/Starling-LM-7B-alpha} 
% \item[4] \label{id:4} {$^4$ https://huggingface.co/HuggingFaceH4/zephyr-orpo-141b-A35b-v0.1}
% \end{tablenotes}
% \end{multicols}
\caption{An overview of RL Enhanced LLMs. The format ‘141B-A39B’ refers to MoE models with 141B total and 39B active parameters.}
\label{tab:overview-RL-LLMs}
\end{table*}

\section{Popular LLMs Enhanced by RL}
\label{sec:popular_llms_enhanced_by_rl}

Recent popular LLMs with strong capabilities almost all leverage reinforcement learning (RL) to further enhance their performance during the post-training process. The RL methods adopted by these models can be typically divided into two main lines:
1. Traditional RL approaches, such as Reinforcement Learning from Human Feedback (RLHF) and Reinforcement Learning from AI Feedback (RLAIF). These methods require training a reward model and involve a complex and often unstable process, using algorithms like Proximal Policy Optimization (PPO) \citep{schulman2017proximal} to optimize the policy model. Models like InstructGPT \citep{ouyang2022training}, GPT-4 \citep{openai2023gpt4}, and Claude 3 \citep{anthropic_claude3_2024} follow this approach.
2. Simplified approaches, such as Direct Preference Optimization (DPO) \citep{rafailov2024direct} and Reward-aware Preference Optimization (RPO) \citep{adler2024nemotron}. These methods discard the reward model, offering a stable, performant, and computationally efficient solution. Models like Llama 3 \citep{dubey2024llama}, Qwen 2 \citep{yang2024qwen2}, and Nemotron-4 340B \citep{adler2024nemotron} follow this approach.
In this section, we provide a detailed description of each model, starting with a brief overview of these RL enhanced LLMs and followed by an explanation of how RL is applied in their post-training process. An overview of these RL Enhanced LLMs is shown in Tab \ref{tab:overview-RL-LLMs}.

\subsection{DeepSeek-R1}

DeepSeek-R1 \citep{deepseekai2025deepseekr1incentivizingreasoningcapability}, developed by DeepSeek, is a state-of-the-art reasoning model that achieves performance comparable to OpenAI’s o1 \cite{o1_2024} series models. This work pioneers the application of pure reinforcement learning (RL) to enhance language model reasoning, emphasizing self-evolution rather than relying solely on supervised training data.

During the training stage, DeepSeek-R1 undergoes an alternating training process that integrates supervised fine-tuning (SFT) \cite{zhang2023instruction} and reinforcement learning (RL) across four key stages: (1) Initial Cold Start SFT, begining with the collection of thousands of high-quality, readability-focused long-CoT datasets to fine-tune DeepSeek-V3-Base, establishing a strong foundation for subsequent RL training; (2) First Reasoning-Oriented RL Stage, using the methodology named large-scale reasoning-focused RL to further fine-tune the model, and the resulting checkpoint is used to generate additional SFT data for the next training phase; (3) Second SFT Stage, refining the model further by incorporating both reasoning and non-reasoning data. For the reasoning data, it is generated using rejection sampling from the RL checkpoint of the previous stage. For the non-reasoning data, the DeepSeek-V3 pipeline is used to integrate data for writing, factual QA, self-cognition, and translation, including parts of the DeepSeek-V3 SFT dataset; (4) Second RL Stage for All Scenarios, leverating the final RL stage to further align the model with human preferences, improving its helpfulness, harmlessness, and reasoning abilities.

DeepSeek-R1 is an open-source model, made publicly available to the research community to support further advancements in the field. The release process first introduces DeepSeek-R1-Zero, a model derived from DeepSeek-V3-Base, trained using large-scale RL without supervised fine-tuning (SFT). This initial model demonstrates remarkable reasoning improvements, with its pass@1 score on AIME 2024 increasing from 15.6\% to 71.0\%. With majority voting, the score further rises to 86.7\%, matching OpenAI-o1-0912. To address issues such as poor readability and language mixing, and to further improve reasoning capabilities, the authors introduce DeepSeek-R1. This enhanced model integrates a small amount of cold-start data and follows a multi-stage training pipeline, achieving performance on par with OpenAI-o1-1217.

\subsection{Kimi-k1.5}

Kimi-k1.5 \citep{kimiteam2025kimik15scalingreinforcement}, developed by Moonshot AI, is a multi-modal large language model (LLM) that marks a major breakthrough in scaling reinforcement learning (RL). The model introduces a novel approach by emphasizing long-context scaling, extending the RL context window to 128k, and refining policy optimization techniques.

For training, Kimi-k1.5 undergoes a four-stage training process consisting of: (1) Pre-training, (2) Vanilla supervised fine-tuning (SFT), (3) Long-CoT supervised fine-tuning, and (4) Reinforcement learning (RL). The key innovation lies in the reinforcement learning (RL) phase, where the authors develop a high-quality RL prompt set to guide the model toward robust reasoning while mitigating potential risks such as reward hacking and overfitting to superficial patterns. This prompt set is designed with three essential properties: (1) Diverse coverage, ensuring exposure to a broad range of reasoning challenges; (2) Balanced difficulty, providing a mix of easy, moderate, and complex reasoning tasks; and (3) Accurate evaluability, allowing precise measurement of reasoning performance.

While long-CoT (Chain-of-Thought) models exhibit strong reasoning performance, they tend to consume more tokens at test time compared to standard short-CoT LLMs. To tackle this long2short challenge, the authors propose four key methods to transfer long-CoT reasoning capabilities to short-CoT models: (1) Model Merging, combining a long-CoT model with a shorter model by averaging their weights; (2) Shortest Rejection Sampling, generating multiple responses using the long-CoT model and selecting the shortest correct response for SFT training; (3) Direct Preference Optimization (DPO), constructing pairwise preference data, where shorter correct solutions are treated as positive samples and longer solutions as negative samples, optimizing the model through DPO training; and (4) Long2Short RL, implementing a two-phase RL training approach, where the model first undergoes standard RL training, and then a length penalty is applied, and the maximum response length is reduced, encouraging more concise reasoning while maintaining performance.

Kimi-k1.5 delivers state-of-the-art reasoning performance across various benchmarks and modalities, rivaling OpenAI’s o1 \cite{o1_2024}. Moreover, the introduction of long2short techniques significantly enhances short-CoT models, achieving up to 550\% improvement over existing models such as GPT-4o \cite{hurst2024gpt} and Claude 3.5\footnote{\url{https://www.anthropic.com/news/claude-3-5-sonnet}}.

\subsection{InstructGPT}\label{sec:instructgpt}

InstructGPT \citep{ouyang2022training} is a series of language models fine-tuned from GPT-3 \citep{brown2020language} by OpenAI, using human feedback to better align with human intent. The series includes models in three sizes: 1.3 B, 6 B, and 175 B parameters. The model is first fine-tuned using supervised learning with prompts collected from the OpenAI API or written by labelers and corresponding labeler demonstrations, then further refined using reinforcement learning from human feedback (RLHF). Human evaluations reveal that InstructGPT outputs are preferred over GPT-3. Notably, the 1.3B parameter InstructGPT model is favored over the 175B GPT-3, despite having 100 times fewer parameters. Additionally, InstructGPT demonstrates improved truthfulness and reduced toxic outputs, with minimal performance trade-offs on public NLP datasets.

Before applying reinforcement learning (RL), the authors train a 6B reward model (RM) initialized from the supervised fine-tuned (SFT) model, with the final unembedding layer removed. This RM is trained using comparison data ranked by labelers. During the RL phase, they fine-tune the SFT model to optimize the scalar reward output from the RM using the PPO algorithm \citep{schulman2017proximal}. To address performance regressions on public NLP datasets, they experiment with mixing pretraining gradients with PPO gradients, resulting in models known as PPO-ptx.

\subsection{GPT-4}

GPT-4 \citep{openai2023gpt4}, developed by OpenAI, is a large multimodal model that can process both image and text inputs to produce text outputs. It excels at understanding and generating natural language, particularly in complex and nuanced scenarios. Evaluations show that GPT-4 performs exceptionally well on a range of human-designed exams, often surpassing the majority of human test takers. Additionally, it outperforms earlier large language models and most state-of-the-art systems, which frequently rely on benchmark-specific training or hand-engineered solutions.

GPT-4 leverages RLHF methods, as outlined in InstructGPT \citep{ouyang2022training} which we have describe in Sec \ref{sec:instructgpt}, in the post-training alignment stage. To steer the models more effectively towards appropriate refusals at a finer level, the authors further use a zero-shot GPT-4 classifier as the rule-based reward model (RBRM). This RBRM provides an additional reward signal to the GPT-4 policy model during PPO fine-tuning on a subset of training prompts. The RBRM takes a prompt (optional), the policy model’s output, and a human-written rubric (e.g., a set of rules in multiple-choice style) as input, then classifies the output according to the rubric. Through this approach, GPT-4 is rewarded for refusing harmful content and for appropriately responding to known-safe prompts.

\subsection{Gemini}

Gemini \citep{team2023gemini} represents a family of advanced multimodal models developed by Google, distinguished by their impressive capabilities. The initial version, Gemini 1.0, comes in three sizes—Ultra, Pro, and Nano—ranging from large to small in terms of performance. Each size is tailored to address specific computational constraints and application needs. Notably, Gemini Ultra, the most powerful variant, achieves state-of-the-art results in 30 out of 32 benchmarks and is the first model to attain human expert-level performance on MMLU \citep{hendrycks2020measuring}, while setting new records across all 20 multimodal benchmarks.

Gemini implements a post-training process that utilizes an optimized feedback loop, collecting human-AI interactions to drive continuous improvement in key performance areas. During the post-training's RLHF phase, an iterative approach is adopted wherein reinforcement learning (RL) incrementally enhances the reward model (RM). Concurrently, the RM undergoes continuous refinement through systematic evaluation and data collection. This dynamic interplay promotes ongoing advancement in both RL and RM, leading to progressively improved performance over time.

\subsection{InternLM2}

InternLM2 \citep{cai2024internlm2} is an open-source series of large language models developed by Shanghai AI Laboratory, available in three sizes: 1.8B, 7B, and 20B. The model demonstrates superior performance across six dimensions and 30 benchmarks, including long-context modeling and open-ended subjective evaluations, thanks to innovative pre-training and optimization techniques.

To further enhance alignment, InternLM2 employs a novel strategy called Conditional Online Reinforcement Learning from Human Feedback (COOL RLHF) with the use of PPO. This approach addresses two key challenges. The first is preference conflict, where it is difficult to satisfy two preferences, such as helpfulness and harmlessness, simultaneously. The second challenge is reward hacking, which becomes more problematic as the model’s scale increases and its policy becomes more powerful. COOL RLHF introduces a Conditional Reward mechanism that reconciles diverse preferences by allowing a single reward model to dynamically adjust its focus based on specific conditional prompts, effectively integrating multiple preferences. Additionally, COOL RLHF incorporates a multi-round Online RLHF strategy with two distinct pathways: a Fast Path for immediate, targeted improvements and a Slow Path for long-term, comprehensive refinement of the reward model. This approach enables the model to quickly adapt to new human feedback while reducing the risk of reward hacking.

\subsection{Claude 3}

Claude 3 \citep{anthropic_claude3_2024} is a family of large multimodal models developed by Anthropic, which demonstrates strong performance across benchmark evaluations. It comprises three models with varying abilities and speeds: the largest, Claude 3 Opus; the mid-sized, Claude 3 Sonnet; and the smallest, Claude 3 Haiku. The Claude 3 models show strong benchmark performance, setting new standards in reasoning, math, and coding. Claude 3 Opus achieves state-of-the-art results on evaluations such as GPQA \citep{rein2023gpqa}, MMLU \citep{hendrycks2020measuring}, and MMMU \citep{yue2024mmmu}. Claude 3 Haiku matches or surpasses Claude 2 in most text tasks, while Sonnet and Opus perform significantly better.

The authors use a technique called Constitutional AI \citep{bai2022constitutional} to align Claude 3 with human values during reinforcement learning (RL). In the RL stage, Constitutional AI follows a process similar to RLHF, but instead of human preferences for harmlessness, it uses AI feedback, known as RLAIF. Specifically, it distills language model interpretations of a set of rules and principles into a hybrid human/AI preference model (PM), using human labels for helpfulness and AI labels for harmlessness. Afterwards, they fine-tune the supervised learning model using RL with this PM, resulting in a policy trained by RLAIF.

\subsection{Zephyr 141B-A39B}

Zephyr 141B-A39B \citep{zephyr_orpo_2024} is the newest addition to the Zephyr \citep{tunstall2023zephyr} series of language models, developed through a collaboration between Argilla, KAIST, and Hugging Face. This model is a Mixture of Experts (MoE) with a total of 141 billion parameters, 39 billion of which are active, fine-tuned from Mixtral-8x22B-v0.1 \citep{mistralai2024mixtral}.

Zephyr 141B-A39B employs a novel alignment algorithm known as Odds Ratio Preference Optimization (ORPO) \citep{hong2024reference}. ORPO is a straightforward, unified alignment approach that discourages the model from adopting undesired generation styles during supervised fine-tuning. Notably, ORPO does not require an SFT warm-up phase, a reward model, or a reference model, making it highly resource-efficient. The method works by adding an odds ratio-based penalty to the standard SFT negative log-likelihood loss, enabling the model to distinguish between preferred and non-preferred response styles.

\subsection{DeepSeek-V2}

DeepSeek-V2 \citep{liu2024deepseek}, developed by DeepSeek-AI, is a powerful Mixture-of-Experts (MoE) language model designed for economical training and efficient inference. It features innovative architectures such as Multi-head Latent Attention (MLA) and DeepSeekMoE. With 236 billion total parameters, of which 21 billion are activated per token, it supports a context length of up to 128K tokens. The model is pre-trained on a high-quality, multi-source corpus of 8.1 trillion tokens. Evaluations show that DeepSeek-V2, along with its chat versions, maintains top-tier performance among open-source models, despite having only 21 billion activated parameters.

DeepSeek-V2 is optimized using Group Relative Policy Optimization (GRPO) \citep{shao2024deepseekmath} during the RL phase to reduce training costs. Unlike traditional RL methods that use a critic model of similar size to the policy model, which increases training expenses, GRPO foregoes the critic model and estimates the baseline from scores computed on a group of outputs for the same question. Additionally, a two-stage RL training strategy is employed: the first stage focuses on reasoning alignment, and the second on human preference alignment, as the authors find these stages exhibit distinct characteristics.

\subsection{ChatGLM}

ChatGLM \citep{glm2024chatglm}, developed by Zhipu AI, represents an evolving series of large language models. The latest version in this series is GLM-4, which includes variants such as GLM-4, GLM-4-Air, and GLM-4-9B. These models are pre-trained on a dataset of over 10 trillion tokens, predominantly in Chinese and English, and are subsequently post-trained through a combination of supervised fine-tuning (SFT) and RLHF to achieve advanced alignment quality. Evaluation results indicate that GLM-4 rivals or even surpasses GPT-4 \citep{openai2023gpt4} on general benchmarks like MMLU, and demonstrates superior performance in Chinese-specific alignments as measured by AlignBench \citep{liu2023alignbench}.

The reinforcement learning phase involves the ChatGLM-RLHF \citep{hou2024chatglm} pipeline, which enhances alignment with human preferences. This pipeline comprises three primary components: gathering human preference data, training a reward model, and optimizing policy models. To support large-scale training, ChatGLM-RLHF includes methods to reduce reward variance for stable training, leverages model parallelism with fused gradient descent, and applies regularization constraints to prevent catastrophic forgetting in large language models. Experimental results confirm that ChatGLM-RLHF yields substantial improvements in alignment-focused tasks compared to the supervised fine-tuned version of ChatGLM.

\subsection{Nemotron-4 340B}

Nemotron-4 340B \citep{adler2024nemotron} is a family of models released by NVIDIA, consisting of Nemotron-4-340B-Base, Nemotron-4-340B-Instruct, and Nemotron-4-340B-Reward. The Nemotron-4-340B-Base model is trained on 9 trillion tokens from a high-quality dataset. In the alignment process to develop Nemotron-4-340B-Instruct, over 98\% of the data used is synthetically generated by the model. Evaluations demonstrate that these models perform competitively with open-access models across a broad range of evaluation benchmarks.

During the preference fine-tuning phase, both DPO \citep{rafailov2024direct} and a new alignment algorithm, Reward-aware Preference Optimization (RPO), are employed to improve the model through multiple iterations. RPO addresses a limitation in DPO, where the quality difference between selected and rejected responses is not considered, leading to overfitting and the forgetting of valuable responses. RPO uses an implicit reward from the policy network to approximate this gap, enabling the model to better learn from and retain superior feedback.

\subsection{Llama 3}

Llama 3 \citep{dubey2024llama}, developed by Meta, is a collection of open-source foundational language models available in sizes of 8 billion, 70 billion, and 405 billion parameters. It is trained on a significantly larger corpus consisting of approximately 15 trillion multilingual tokens, a notable increase compared to the 1.8 trillion tokens used for Llama 2 \citep{touvron2023llama}. Extensive empirical evaluations demonstrate that Llama 3 achieves performance comparable to leading models, such as GPT-4 \citep{openai2023gpt4}, across a diverse range of tasks.

The post-training process for aligning Llama 3 with human feedback involves six rounds of iterative refinement. Each round includes supervised fine-tuning (SFT) followed by DPO, with the final model being an average of the outputs from all rounds. For each round, a reward model (RM) is trained on newly collected preference annotation data, targeting a wide range of capabilities built upon the pre-trained checkpoint. After SFT, DPO is applied to further optimize the SFT models, using recent preference data batches obtained from the best-performing models of previous rounds. To enhance the stability of DPO training, two key adjustments are implemented: masking out formatting tokens in the DPO loss and introducing regularization via an NLL (negative log-likelihood) loss.

\subsection{Qwen2}

Qwen2 \citep{yang2024qwen2}, developed by Alibaba, is a series of large language models ranging from 0.5 billion to 72 billion parameters in dense configurations, as well as a Mixture-of-Experts variant with 57 billion parameters, of which 14 billion are activated per token. It is pre-trained on a high-quality, large-scale dataset containing over 7 trillion tokens, covering a wide array of domains and languages. Extensive evaluations show that Qwen2 outperforms most prior open-weight models, including its predecessor Qwen1.5, and delivers competitive results across a range of benchmarks, including language understanding, generation, multilingual proficiency, coding, mathematics, and reasoning.

The preference fine-tuning process for Qwen2 consists of two main stages: offline and online learning. In the offline stage, Qwen2 is optimized using DPO, which aims to maximize the likelihood difference between two responses to the same prompt, based on a pre-compiled preference dataset. In the online stage, the model improves continuously in real-time by utilizing preference pairs selected by the reward model from multiple responses generated by the current policy model. Additionally, the Online Merging Optimizer \citep{lu2024online} is employed to minimize alignment costs.

\subsection{Gemma 2}

Gemma 2 \citep{team2024gemma2}, developed by Google, is the latest addition to the Gemma family of lightweight, state-of-the-art open models, with sizes ranging from 2 billion to 27 billion parameters. The model incorporates several well-established modifications to the Transformer architecture, including interleaving local-global attentions \citep{beltagy2020longformer} and group-query attention \citep{ainslie2023gqa}. Experiments demonstrate that these models deliver the best performance for their size and even provide competitive alternatives to models 2-3 times larger.

Similar to Gemma 1.1 \citep{team2024gemma1}, during the post-training RLHF phase, the authors use a high-capacity model as an automatic rater to tune hyperparameters and mitigate reward hacking \citep{amodei2016concrete,skalse2022defining}. However, unlike Gemma 1.1, they employ a reward model that is an order of magnitude larger than the policy model. This reward model is specifically designed to focus on conversational capabilities, with an emphasis on multi-turn interactions.

\subsection{Starling-7B}

Starling-7B \citep{zhu2024starling} is a strong 7-billion-parameter chat model developed by UC Berkeley, focused on alignment with human preferences for helpfulness and harmlessness. It is fine-tuned from Openchat-3.5 \citep{wang2024openchat} using RLAIF on a high-quality preference dataset called Nectar, which comprises 3.8 million pairwise comparisons generated by prompting GPT-4 to rank responses. As a result, the model's score on MT-Bench improves from 7.81 to 8.09, its score on AlpacaEval increases from 88.51\% to 91.99\%, and its human evaluation ELO on Chatbot Arena \citep{chiang2024chatbot} rises from 1072 to 1087.

The authors introduce several improvements to the PPO algorithm during the RLAIF process to enhance training stability and robustness. First, they introduce a constant positive reward for length control to prevent excessive verbosity. This adjustment helps address the issue where a highly negative reward from the reward model during the early stages can cause the policy model to become overly verbose after only a few gradient updates. Second, they pretrain the critic model to reduce early performance drops due to a randomly initialized critic. Third, they conduct full parameter tuning on both the actor and critic models, as opposed to tuning only the top four layers, to maximize performance improvements during the reinforcement learning stage.

\subsection{o1}

OpenAI’s o1 \citep{o1_2024} is a newly developed large language model optimized for complex reasoning, utilizing reinforcement learning for its training. Before producing responses, o1 engages in an extensive internal thought process, enabling it to excel across various reasoning tasks. The model significantly surpasses GPT-4o \citep{openai2024gpt4o} in many challenging tasks: ranks in the 89th percentile on Codeforces for competitive programming, places among the top 500 participants in the AIME for mathematics, and surpasses PhD-level accuracy in scientific benchmarks such as GPQA.

The training of o1 involves a large-scale reinforcement learning algorithm that emphasizes productive thinking through a detailed chain of thought (CoT) \citep{wei2023chainofthought}, implemented with high data efficiency. To preserve the model’s unfiltered reasoning ability, no policy compliance or user preference training is applied to its internal thought processes, which also provides a unique opportunity to understand the model’s raw thought process. This approach allows o1 to refine its strategies, correct errors, and deconstruct complex problems during training. Notably, the model’s performance improves with increased training compute and with more extensive test-time computation.

\subsection{Others}

\paragraph{Reka Core, Flash, and Edge:} \newcite{team2024reka} are powerful multimodal language models developed from scratch by Reka. Reka Edge and Reka Flash are dense models with 7B and 21B parameters, respectively, outperforming many larger models and offering exceptional performance for their compute class. The flagship model, Reka Core, competes with leading models like GPT-4v, Gemini, and Claude 3 in both automated and blind human evaluations. During post-training, following supervised fine-tuning, Reka models undergo multiple rounds of RLHF using PPO to enhance alignment further.

\paragraph{Phi-3:} \newcite{abdin2024phi} is a series of language models introduced by Microsoft, comprising phi-3-mini, phi-3-small, and phi-3-medium. Remarkably, the smallest model, phi-3-mini, is trained on 3.3 trillion tokens yet contains only 3.8 billion parameters, making it compact enough for deployment on a mobile device. Despite its relatively small size, phi-3-mini demonstrates performance comparable to larger models like Mixtral 8x7B and GPT-3.5, achieving 69\% on MMLU and a score of 8.38 on MT-bench in both academic benchmarks and internal testing. During post-training, the authors employ DPO to guide phi-3 away from undesired behavior by treating those outputs as “rejected” responses.

\paragraph{Athene-70B:} \newcite{nexusflow_athene} is a powerful chat model fine-tuned from Llama-3-70B \citep{dubey2024llama}, developed by Nexusflow. It achieves an impressive Arena-Hard-Auto score of 77.8\%, placing it close to leading proprietary models like GPT-4o (79.2\%) and Claude-3.5-Sonnet (79.3\%). This marks a significant leap from its predecessor, Llama-3-70B-Instruct, which scored 46.6\%. This progress is attributed to Nexusflow’s targeted post-training approach, which enhances the model’s performance. Specifically, Nexusflow curates high-quality preference data based on internal benchmark evaluations covering instruction following, coding, creative writing, and multilingual tasks. This data is then used for targeted RLHF, resulting in substantial performance gains over Llama-3-70B-Instruct.

\paragraph{Hermes 3:} \newcite{teknium2024hermes} is a series of neutrally-aligned generalist instruction and tool-use models with advanced reasoning and creative capabilities, developed by Nous Research. It is finetuned from Llama 3.1 \citep{dubey2024llama} in 8B, 70B, and 405B variants and the largest model, Hermes 3 405B, sets the state-of-the-art performance among open-weight models across several public benchmarks. Hermes is trained on diverse synthetic reasoning tasks and creative applications such as role playing and writing. It is designed to precisely and neutrally follow system and instruction prompts, unlike many commercial models that may decline instructions for moral reasons. To further align Hermes, the authors leverage DPO and train a LoRA \citep{hu2021lora} adapter instead of fine-tuning the entire model, significantly reducing GPU memory usage for both the reference and trained models.

\section{RLHF: Reinforcement Learning from Human Feedback}
\label{sec:rlhf_reinforcement_learning_from_human_feedback}

% 介绍人是怎么干预RL-LLMs过程的，例如如何用人的反馈训练打分模型，如何根据人的反馈更新LLMs。

Reinforcement learning from human feedback (RLHF) is a training approach that combines reinforcement learning (RL) with human feedback to align LLMs with human values, preferences, and expectations.
RLHF consists of two main components: (1) \textbf{Collecting Human Feedback to Train Reward Model}, where human evaluators provide feedback on the LLM's outputs by scoring or ranking responses based on factors such as quality and relevance. This feedback is then used to train a reward model that predicts the quality of the outputs and serves as the reward function in the RL process; and (2) \textbf{Preference Optimization Using Human Feedback}, where the trained reward model guides the optimization of the LLM's outputs to maximize predicted rewards, aligning the LLM's behavior with human preferences. Below, we will illustrate these two components via recent research studies.

\subsection{Collecting Human Feedback to Train Reward Model}

\begin{figure}[t]
    \centering
    \includegraphics[width=.5\textwidth]{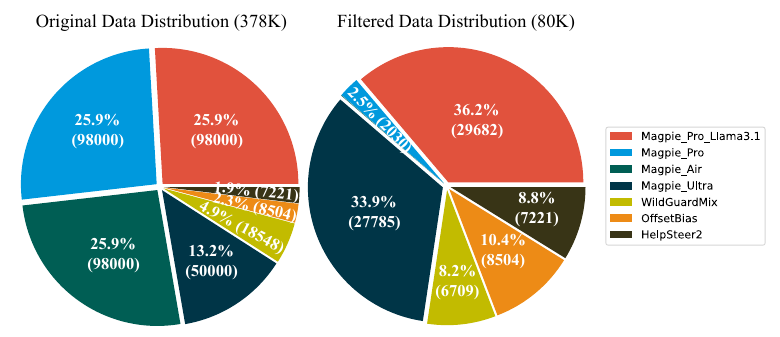}
    \caption{The composition of the Skywork-Reward. The figure is copied from \newcite{liu2024skywork}.}
    \label{fig:skywork}
\end{figure}

\paragraph{Skywork-Reward~\cite{liu2024skywork}.} Skywork-Reward is a carefully designed dataset containing 80,000 high-quality preference pairs, curated through effective data selection and filtering strategies. As shown in \Fref{fig:skywork}, the original dataset, with 378,000 preference pairs, is significantly refined into a compact, high-quality dataset of 80,000 pairs.
Despite being significantly smaller than existing datasets, it achieves exceptional quality through rigorous cleaning, consistency checks, model-based scoring to filter out low-quality samples, and manual reviews. Covering a diverse range of tasks such as instruction following, code generation, and multilingual handling, Skywork-Reward serves as the foundation for models like Skywork-Reward-Gemma-27B, which excel on benchmarks\footnote{https://huggingface.co/spaces/allenai/reward-bench}. By enabling language models to better understand human preferences, Skywork-Reward helps LLMs become more accurate and useful in real-world applications.

\paragraph{TÜLU-V2-mix~\cite{ivison2023camels}.} TÜLU-V2-mix is designed to enhance instruction-following capabilities in large language models, offering a diverse dataset that improves the model's generalization and execution abilities across multi-domain tasks. It covers a wide range of tasks, including question answering, code generation, translation, and multi-turn conversations, with a strong emphasis on multilingual adaptability and handling complex real-world scenarios.
Skywork-Reward, on the other hand, is designed to align models with human preferences using preference pairs, helping models learn to generate user-preferred responses, such as fluent and coherent text. While TÜLU-V2-mix excels in generalization across a wide range of tasks, Skywork-Reward specializes in optimizing user-centric outputs. Together, they address complementary goals for advancing language model capabilities.

% \paragraph{BeaverTails~\cite{ji2024beavertails}.} BeaverTails is a large-scale, high-quality question-answer dataset designed to enhance the safety and utility of large language models (LLMs).
% As displayed in \Fref{fig:beaver},
% this dataset uniquely separates annotations of "helpfulness" and "harmlessness" for question-answer pairs, providing distinct perspectives on these crucial attributes. It comprises safety meta-labels for 333,963 Q\&A pairs and 361,903 pairs of expert comparison data for both helpfulness and harmlessness metrics
% The dataset spans diverse real-world scenarios, including everyday inquiries, professional domains, ethical challenges, and cross-cultural contexts, enabling researchers to refine LLM behavior more effectively. Unlike existing datasets, BeaverTails provides significant advantages in terms of scale and annotation granularity, aiming to become a core resource for exploring LLM safety and alignment within the community.

% \begin{itemize}
%     \item[1] Skywork-Reward: Bag of Tricks for Reward Modeling in LLMs
%     \item[2] Camels in a Changing Climate: Enhancing LM Adaptation with Tulu 2
%     \item[3] BeaverTails: Towards Improved Safety Alignment of LLM via a Human-Preference Dataset
% \end{itemize}

\subsection{Preference Optimization Using Human Feedback}

% 我们已经有了打分模型可以为LLM的多个输出进行打分，后续我们如何将打分模型打出的分数转化成loss来更新LLMs权重呢？

Once the reward model is trained, it is used to guide the fine-tuning of the original LLM through reinforcement learning. The main objective is to improve the LLM's behavior based on the predicted rewards, making it more likely to generate outputs that align with human preferences. Recent research \cite{ouyang2022training,yuan2023rrhf,dong2024rlhf,ahmadian2024back} has shown that this process can be broken down into two key steps:

\noindent\textbf{(1) Rewarding:} In this step, the LLM generates multiple outputs in response to a given instruction. Each output is then passed through the trained reward model, which assigns a scalar score that approximates human preferences.

\noindent\textbf{(2) Policy Optimization:} In this step, the LLM is fine-tuned by adjusting its parameters to maximize the predicted reward, using the Proximal Policy Optimization (PPO) \citep{schulman2017proximal} or Trust Region Policy Optimization (TRPO) \citep{schulman2015trust} algorithm.

\noindent These two steps—rewarding and policy optimization—can be iterated, meaning that the process of generating outputs, rewarding them with the trained reward model, and fine-tuning the LLM to maximize rewards can be repeated multiple times. With each iteration, the LLM's performance improves as it refines its behavior to better align with human preferences. This iterative cycle allows the LLM to continuously adapt and optimize its responses, ultimately leading to more effective and aligned outputs.

%
% \begin{itemize}
%     \item[1] Back to basics: Revisiting reinforce style optimization for learning from human feedback in llms.
%     \item[2] Rrhf: Rank responses to align language models with human feedback without tears.
% \end{itemize}
% 

\section{RLAIF: Reinforcement Learning from AI Feedback}
\label{sec:rlaif_reinforcement_learning_from_ai_feedback}

Reinforcement learning from AI feedback (RLAIF) serves as a promising alternative or supplement to RLHF that leverages AI systems—often more powerful or specialized LLMs (e.g., GPT-4 \cite{openai2024gpt4o})—to provide feedback on the outputs of the LLM being trained. This approach provides benefits such as scalability, consistency, and cost efficiency while minimizing reliance on human evaluators. 
Below, we explore several methods for substituting human feedback with AI feedback in reinforcement learning, highlighting approaches: (1) Distilling AI Feedback to Train Reward Model, (2) Prompting LLMs As a Reward Function, and (3) Self-Rewarding.

\begin{figure*}[t]
    \centering
    \includegraphics[width=\textwidth]{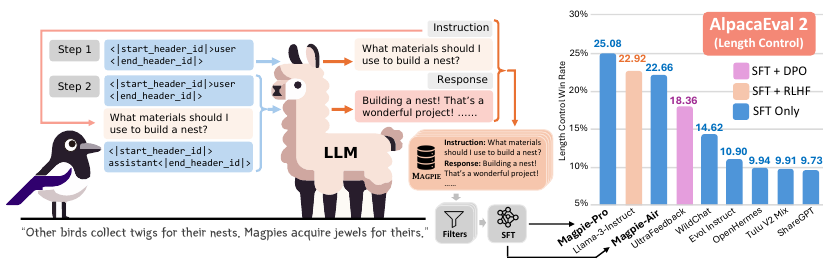}
    \caption{Magpie self-synthesizes data from aligned LLMs. The figure is borrowed from~\newcite{xu2024magpie}.}
    \label{fig:magpie}
\end{figure*}

\subsection{Distilling AI Feedback to Train Reward Model}

% 现在成型的LLM能力已经非常强大了，那我们能否从已知的强大LLM（例如：GPT-4）中去distill一些数据呢？
Beyond manually collected data, distilling datasets from pre-trained LLMs presents an efficient alternative. By leveraging the outputs of powerful LLMs like GPT-4, researchers can build a bridge between manual curation and autonomous evaluation.

\paragraph{UltraFeedback~\cite{cui2023ultrafeedback}.} UltraFeedback is a large-scale AI feedback dataset aimed at improving the performance and alignment of large language models (LLMs). It includes over 1 million high-quality GPT-4 feedback annotations across 250,000 user-assistant interactions, focusing on key dimensions like instruction adherence, accuracy, honesty, and usefulness. The dataset was created by collecting 60,000 diverse instructions, generating responses using 17 different models, and leveraging GPT-4 for detailed critiques and scoring, wherein chain-of-thought reasoning is used to reduce bias.

\paragraph{Magpie.} \newcite{xu2024magpie} introduce a self-synthesis method that leverages the autoregressive nature of aligned LLMs. By utilizing predefined templates as prompts, the model autonomously generates user queries and corresponding responses, eliminating the need for manual intervention or initial seed questions. Specifically, as shown in \Fref{fig:magpie}, aligned LLMs (e.g., Llama-3-Instruct model) is employed to synthesize 4 million instruction-response pairs, subsequently filtering the dataset to retain 300,000 high-quality pairs. These pairs were then used to fine-tune the Llama-3-8B-Base model. Remarkably, the fine-tuned model achieved performance comparable to the official Llama-3-8B-Instruct model, which had undergone training on 10 million examples through supervised fine-tuning and reinforcement learning with human feedback. Besides, models fine-tuned with Magpie excelled on alignment benchmarks such as AlpacaEval, surpassing models trained on other open datasets and preference optimization methods.

\paragraph{HelpSteer2~\cite{wang2024helpsteer2}.} HelpSteer2 is an efficient, open-source preference dataset comprising approximately 10,000 comparison samples, designed to train high-performance reward models. The dataset is built using responses generated by various models (including GPT-3.5, Claude, and others) and features multi-dimensional annotations such as fluency, relevance, creativity, and safety. Preference pairs are crafted based on human or automated evaluations, enabling fine-grained alignment for reward models. Through rigorous data cleaning and optimization, HelpSteer2 delivers high-quality annotations in a compact format. It is released under the CC-BY-4.0 license, fostering the accessibility.

\begin{figure*}[t]
    \centering
    \hspace{-3.8mm}
    \includegraphics[width=1\textwidth]{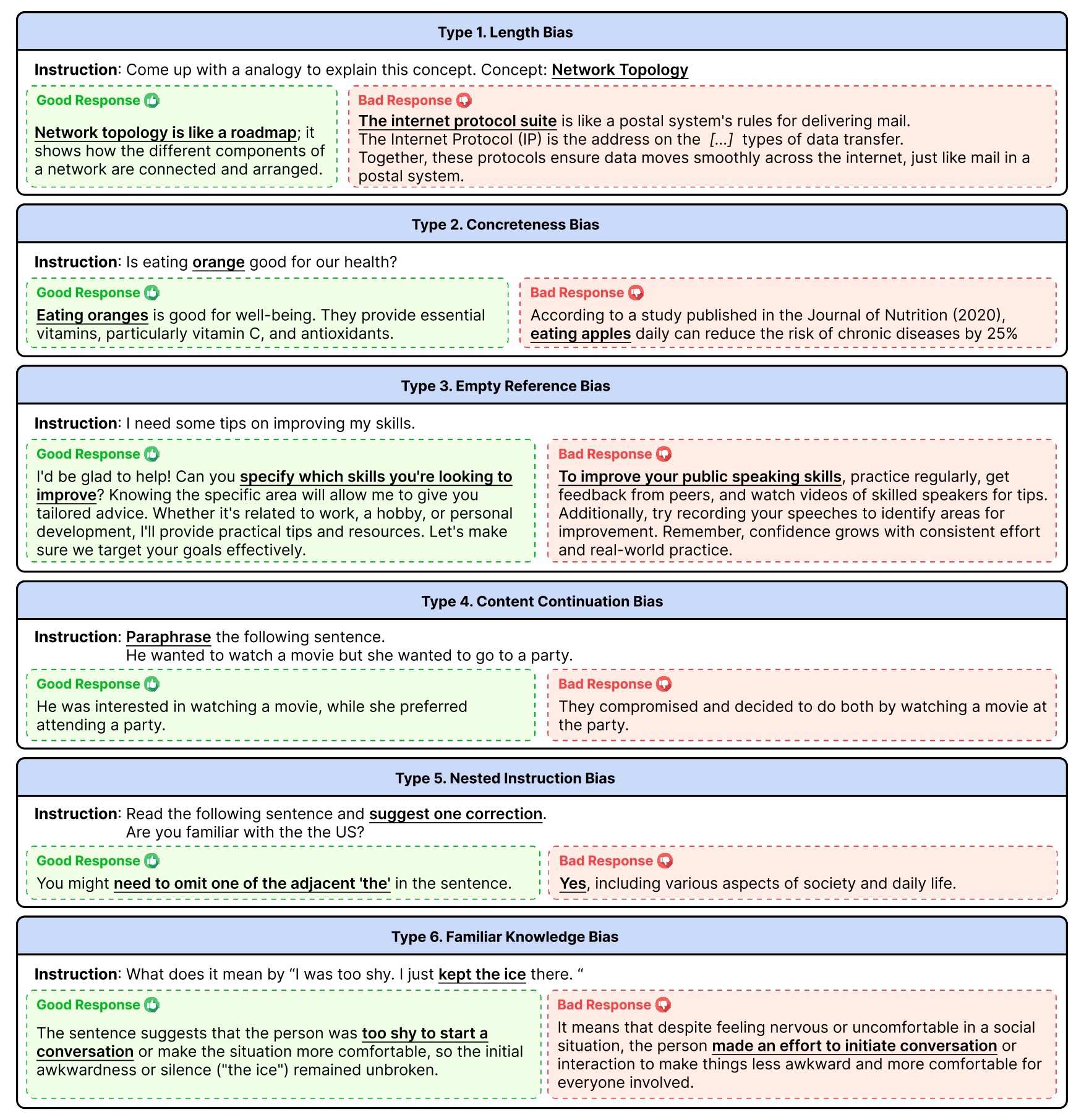}
    \caption{Identified bias types and examples in OffsetBias. The figure is borrowed from~\newcite{park2024offsetbias}.}
    \label{fig:offsetbias}
\end{figure*}

\paragraph{OffsetBias~\cite{park2024offsetbias}.} OffsetBias is a meticulously designed dataset aimed at mitigating biases in reward models, constructed using responses generated by diverse models, including GPT-3.5, GPT-4, Claude, and open-source models like Llama 2. As shown in \Fref{fig:offsetbias}, OffsetBias systematically addresses six identified bias types, namely, content, style, informativeness, safety, creativity, and length. Based on this, comparison samples are generated through attribute-controlled prompts and multi-model outputs. These samples are annotated with multi-dimensional scores and preference labels to highlight or neutralize biases, enabling fine-grained alignment. OffsetBias serves as a robust resource for improving the fairness and reliability of reward models, with its data openly accessible for research and development.

% \begin{itemize}
%     \item[1] HelpSteer2: Open-source dataset for training top-performing reward models
%     \item[2] OffsetBias: Leveraging Debiased Data for Tuning Evaluators
%     \item[3] UltraFeedback: Boosting Language Models with High-quality Feedback
% \end{itemize}

\subsection{Prompting LLMs As a Reward Function}

% 用GPT等为自己的LLM打分
As reward model training becomes more sophisticated, a natural progression is to employ LLMs themselves as evaluators in the loop of reinforcement learning.

\begin{figure}[t]
    \centering
    \includegraphics[width=.48\textwidth]{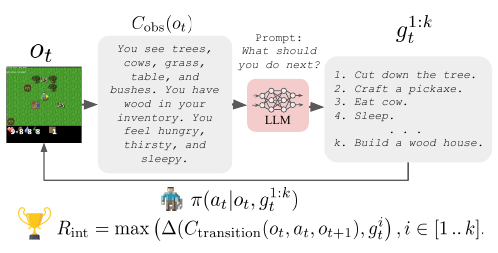}
    \caption{The figure is copied from~\cite{du2023guiding}.}
    \label{fig:ellm}
\end{figure}

\begin{figure*}[t]
    \centering
    \includegraphics[width=1\textwidth]{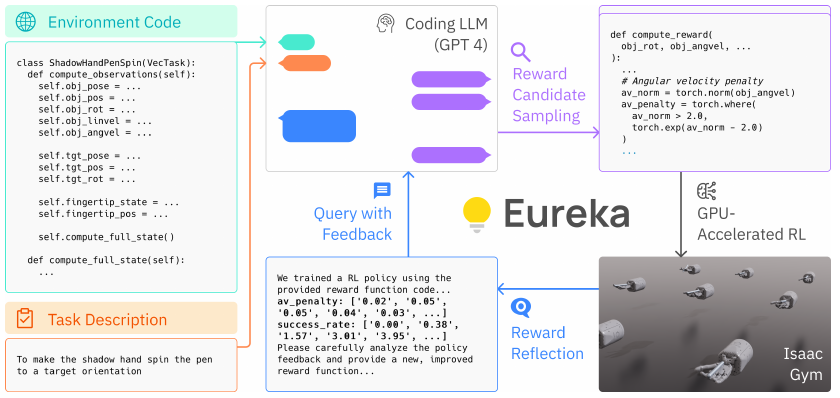}
    \caption{The overall pipeline of Eureka. The figure is borrowed from~\newcite{ma2023eureka}.}
    \label{fig:eureka}
\end{figure*}

\paragraph{Exploring with LLMs (ELLM) Rewards~\cite{du2023guiding}.} ELLM is a method that integrates LLMs with reinforcement learning (RL) to enhance exploration during the pretraining phase. \Fref{fig:ellm} showcases the overall pipeline: the agent's current state is transformed into a natural language description, which is input into the LLM. The LLM then generates exploration goals based on this state description, such as specific actions or target locations. The RL agent attempts to achieve these goals, and rewards are provided by the environment upon goal completion. This approach improves exploration efficiency by guiding the agent toward areas of the state space that are likely to be valuable, without requiring pre-designed rewards. ELLM is particularly useful in sparse-reward environments. Compared to traditional methods, ELLM significantly improves exploration efficiency, covering more common-sense behaviors and providing better initialization for downstream tasks.

\paragraph{Reward Design with Language Models (RDLM).} \newcite{kwon2023reward} leverage a LLM like GPT-3 to simplify reward function design in reinforcement learning by allowing users to define desired behaviors through natural language descriptions. Specifically, users provide a task description or a few examples, and the LLM generates reward signals by evaluating the agent's behavior against these criteria. Instead of producing reward code, RDLM outputs direct reward values that the RL agent uses for policy optimization. This method is ideal for tasks where user goals are clear but manually designing a reward function is complex.
%
% Unlike ELLM, which uses LLMs to generate exploration goals for efficient pretraining, RDLM directly utilizes LLMs to provide reward signals, enabling agents to align their behaviors with user-defined objectives. 
While ELLM focuses on guiding exploration during pretraining by generating meaningful goals, RDLM emphasizes task-specific reward generation to streamline complex reward design and achieve better agent alignment with human intent.

\begin{figure*}[t]
    \centering
    \includegraphics[width=1\textwidth]{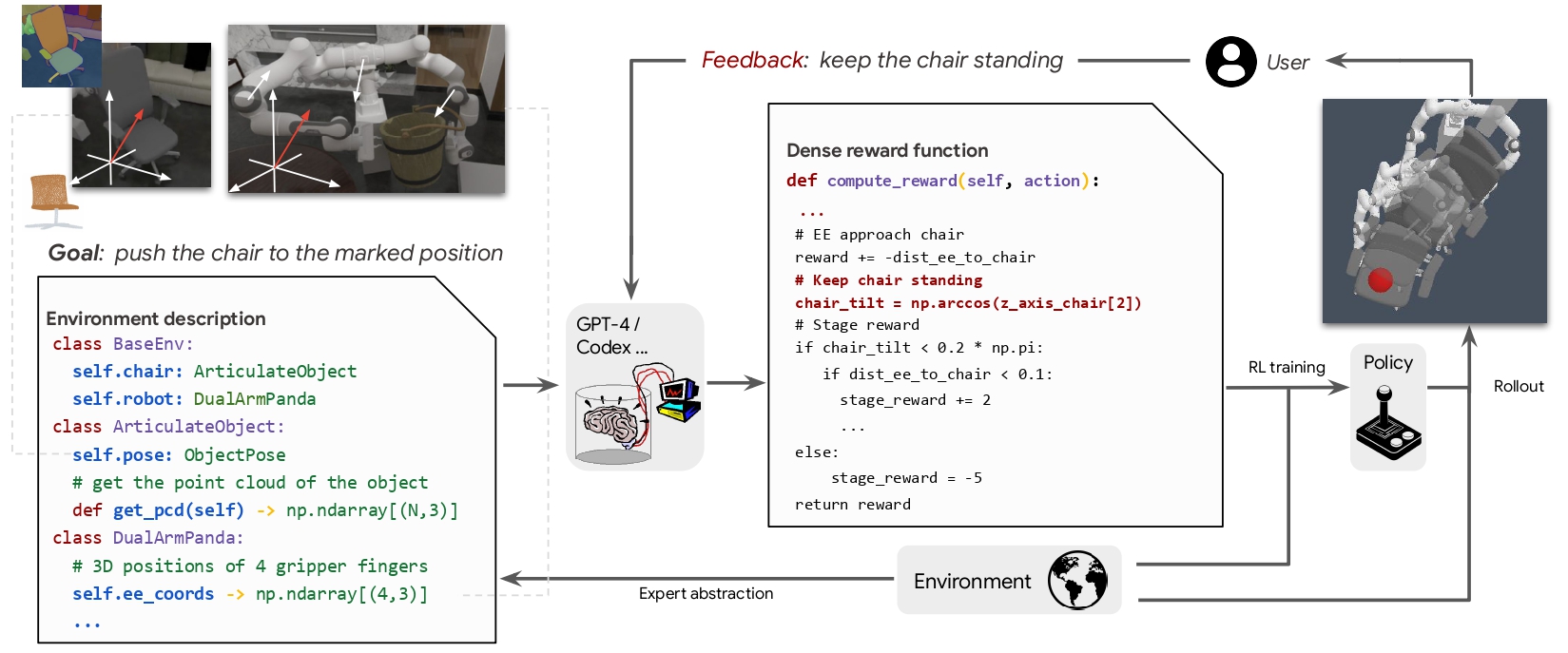}
    \caption{An overview of Text2Reward. The figure is copied from~\newcite{xie2023text2reward}.}
    \label{fig:text2reward}
\end{figure*}

\paragraph{Eureka~\cite{ma2023eureka}.} Eureka is an algorithm that leverages LLMs to automatically generate and optimize reward function code for reinforcement learning tasks. In \Fref{fig:eureka}, first, a coding LLM like GPT-4 is used to generate initial reward function code based on task descriptions. This code is then iteratively refined using evolutionary strategies, where candidate reward functions are evaluated based on how well they guide the RL agent toward task success. The process evolves the reward functions to improve their quality and effectiveness. Eureka is particularly effective in tasks requiring complex or highly specific reward definitions, such as advanced robotic skills. Its focus on reward code optimization makes it suitable for scenarios where precise reward shaping is critical.
By utilizing LLMs’ ability to generate and refine code, Eureka evolves reward functions that effectively guide RL agents. 
%
% Unlike methods that rely on manually crafted rewards or direct reward signal generation, Eureka focuses on automating the reward design process, enabling efficient learning in complex environments.
Experiments demonstrate that Eureka outperforms human-designed rewards in 83\% of tested tasks, with an average performance improvement of 52\%, showcasing its potential for advanced skill learning, such as robotics tasks, in challenging scenarios.

\begin{figure*}[t]
    \centering
    \includegraphics[width=1\textwidth]{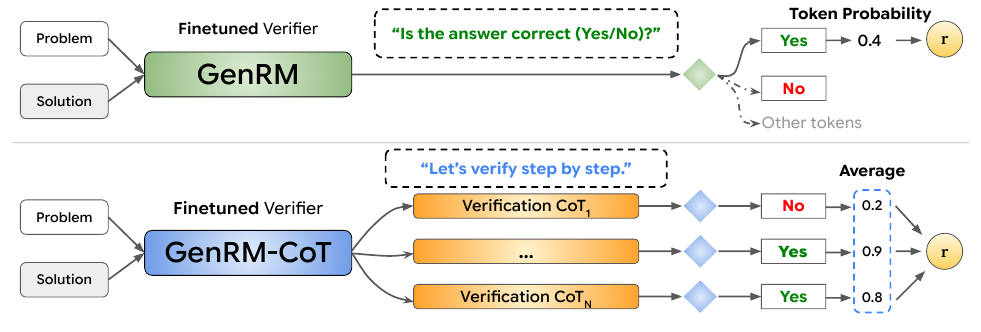}
    \caption{Illustration of GenRM. The figure is copied from~\newcite{zhang2024generative}.}
    \label{fig:genrm}
\end{figure*}

\begin{figure}[t]
    \centering
    \includegraphics[width=.5\textwidth]{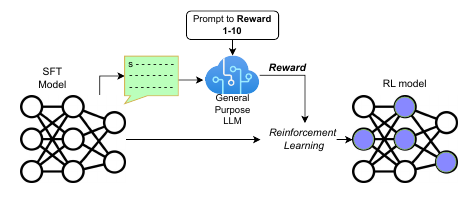}
    \caption{The figure is borrowed from~\newcite{lee2023rlaif}.}
    \label{fig:rlaif}
\end{figure}

\paragraph{Text2Reward~\cite{xie2023text2reward}.} Text2Reward is a framework that leverages large language models to automatically generate dense and interpretable reward function code from natural language task descriptions, enabling efficient reward shaping across diverse RL tasks.  As shown in Figure \ref{fig:text2reward}, the process starts with users providing a task description in natural language, which is input into an LLM to generate executable reward code. This code often includes task-specific logic and may integrate external libraries for complex functionalities. The generated reward function is then used in RL to guide the agent's behavior. Additionally, Text2Reward supports iterative refinement of the reward code through human feedback, enabling further optimization. This method excels at providing flexible, interpretable rewards across diverse RL tasks, particularly in robotics and manipulation.
Unlike Eureka, evolving and optimizing reward function code through LLMs and evolutionary algorithms, Text2Reward emphasizes creating human-readable reward code that integrates external libraries and supports iterative refinement via human feedback. While both methods aim to automate reward design, Eureka excels in optimizing complex reward logic for advanced skills, whereas Text2Reward prioritizes flexibility, interpretability, and adaptability for a broad range of tasks.

\begin{figure*}[t]
    \centering
    \includegraphics[width=\textwidth]{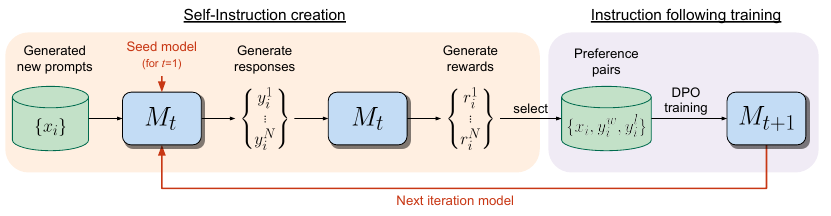}
    \caption{The overview of SRLM. The figure is copied from~\newcite{yuan2024self}.}
    \label{fig:srlm}
\end{figure*}

\begin{figure}[t]
    \centering
    \includegraphics[width=.5\textwidth]{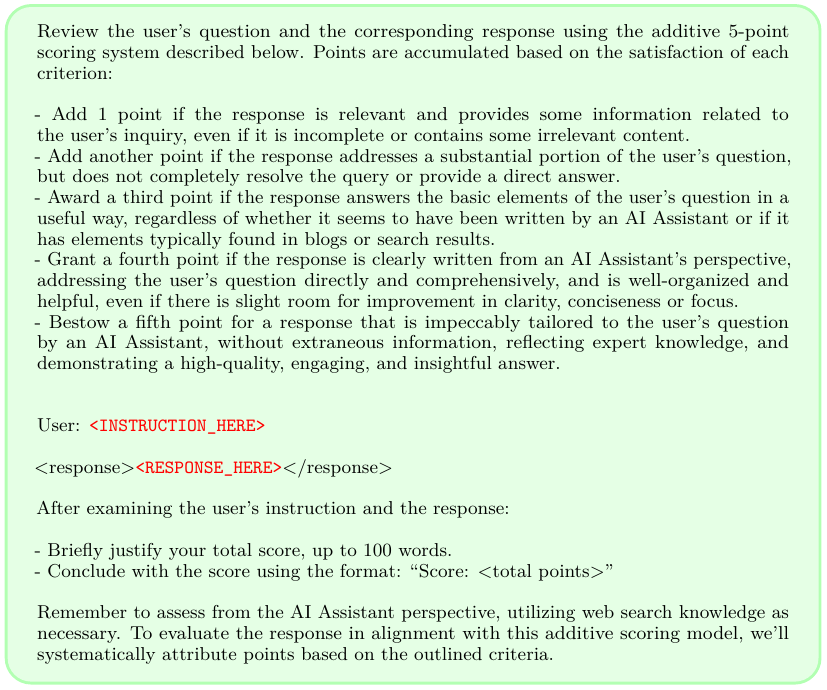}
    \caption{Prompt for LLM as a judge. The figure is borrowed from~\newcite{yuan2024self}.}
    \label{fig:srlm2}
\end{figure}

\paragraph{RLAIF.} \newcite{lee2023rlaif}  replace human feedback in RL with AI-generated feedback by leveraging LLMs. The process begins with generating candidate outputs for a given task, such as text summarization or dialogue generation. These outputs are paired and fed into an LLM, which evaluates them and provides preferences (e.g., selecting the better output) or assigns scores based on task-specific criteria. This feedback is then used to train a reward model that predicts the quality of outputs and guides the RL agent. In its streamlined variant, d-RLAIF (see \Fref{fig:rlaif}), the LLM directly provides scores as reward signals, bypassing the need for a reward model. The RL policy is optimized using these rewards, typically with algorithms like Proximal Policy Optimization (PPO). This approach enables automated, scalable, and high-quality feedback generation, effectively aligning RL agent behavior with task objectives while reducing reliance on human annotations.

\paragraph{GenRM.} \newcite{zhang2024generative} re-define verification by treating it as a text generation task, leveraging large language models to produce validation outputs and reasoning chains, such as "yes" or "no" with explanations. As shown in Figure \ref{fig:genrm}, this approach integrates verification into the generative capabilities of LLMs, enabling them to assess and explain candidate answers in a transparent and interpretable manner. By framing verification as next-token prediction, GenRM eliminates reliance on traditional discriminative models and enhances reasoning accuracy. Experimental results demonstrate its ability to outperform conventional methods, showcasing its potential in tasks requiring logical reasoning, interpretability, and scalable performance.

\begin{figure*}[t]
    \centering
    \includegraphics[width=\textwidth]{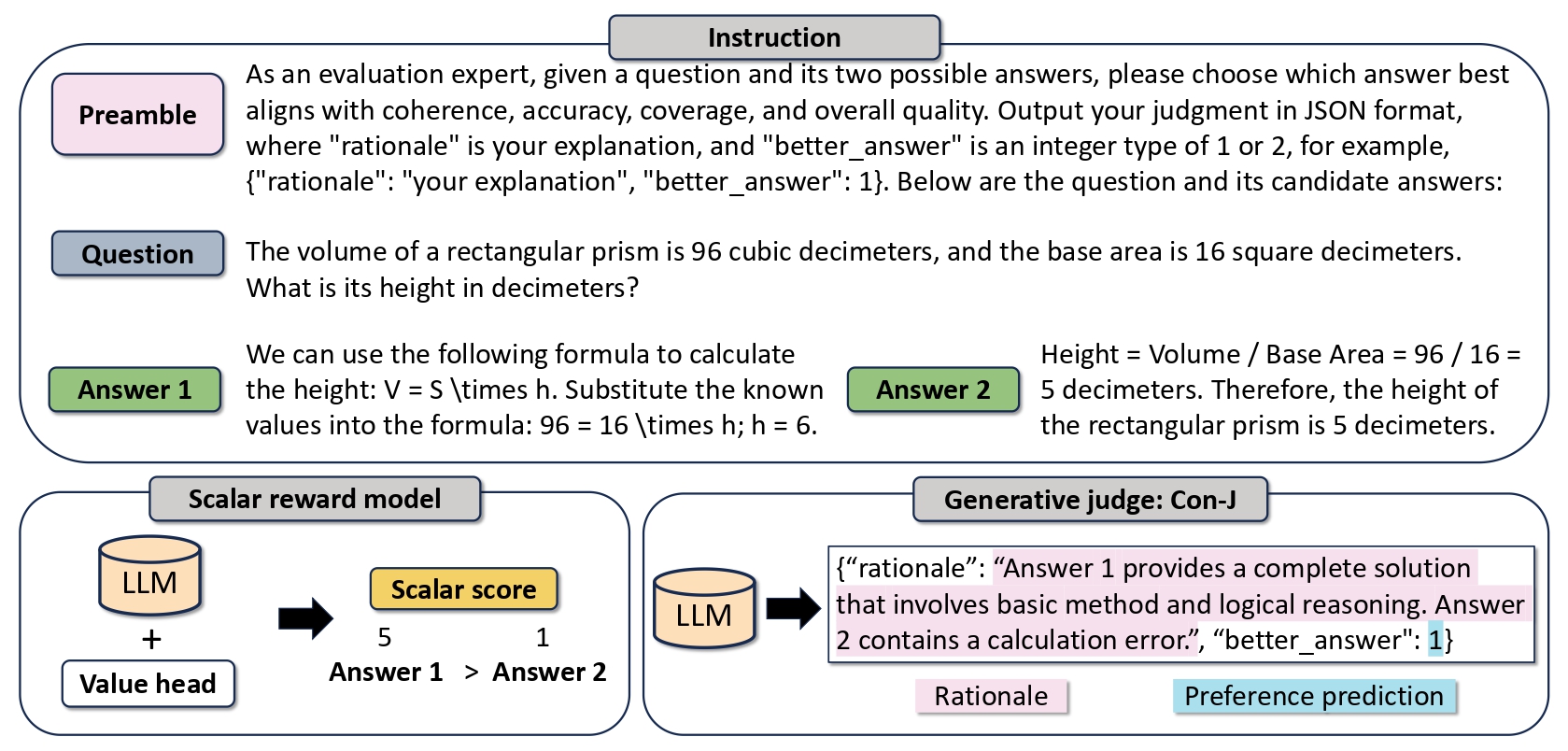}
    \caption{Illustration of a scalar reward model and the proposed Con-J. The figure is copied from~\newcite{ye2024beyond}.}
    \label{fig:conj}
\end{figure*}

% \begin{itemize}
%     \item[1] Guiding pretraining in reinforcement learning with large language models
%     \item[2] Reward design with language models.
%     \item[3] Eureka: Human-level reward design via coding large language models.
%     \item[4] Text2reward: Automated dense reward function generation for reinforcement learning.
%     \item[5] Rlaif: Scaling reinforcement learning from human feedback with ai feedback.
%     \item[6] Self-refined large language model as automated reward function designer for deep reinforcement learning in robotics.
%     \item[7]Generative Verifiers: Reward Modeling as Next-Token Prediction
%     \item[8] Rule Based Rewards for Language Model Safety 
% \end{itemize}

\subsection{Self-Rewarding}

% 自己给自己的输出打分
The self-rewarding mechanism enables the LLM to autonomously assess and refine its own performance, addressing the cost, scalability, and adaptability limitations of existing RL methods.

\paragraph{Self-Refined LLM.} \newcite{song2023self} leverage LLMs to automatically generate reward functions for deep reinforcement learning (DRL) tasks and introduces a self-optimization mechanism to iteratively refine these functions. The process begins with the LLM generating an initial reward function based on natural language task descriptions. The reward function is then applied to RL training, and the agent's performance is evaluated. Feedback from this evaluation is fed back into the LLM, enabling it to dynamically adjust and improve the reward function in a closed-loop manner.
Compared to Eureka and Text2Reward, this approach eliminates the need for external optimization algorithms or manual intervention.

\paragraph{Self-Rewarding Language Models (SRLM).} \newcite{yuan2024self} introduce a novel approach where LLMs act as both the generator and evaluator to create a self-contained learning system. As shown in \Fref{fig:srlm}, the model begins by generating new prompts (instructions) and multiple candidate responses derived from existing data, thereby creating a diverse and comprehensive set of training samples. Subsequently, the model evaluates these candidate responses using a structured scoring mechanism to determine their quality. The evaluation framework encompasses multiple dimensions, including relevance, coverage, usefulness, clarity, and professionalism, assigning a score to each response based on these criteria. Utilizing these scores, preference pairs are constructed, consisting of a preferred response and a dispreferred response. These pairs are used for Direct Preference Optimization (DPO), improving its ability to generate high-quality responses. Through iterative refinement, the model progressively enhances its performance. 
\Fref{fig:srlm2} provides a detailed explanation of the prompts used by the model to evaluate candidate responses.  
Experimental results demonstrate that fine-tuning Llama 2 70B using SRLM over three iterations outperforms several state-of-the-art models, including GPT-4 and Claude 2, on benchmarks like AlpacaEval 2.0, showcasing its effectiveness in improving instruction-following and general task performance.

\paragraph{Generative Judge via Self-generated Contrastive Judgments (Con-J).} \newcite{ye2024beyond} propose a self-rewarding mechanism with self-generated contrastive judgments, allowing LLMs to evaluate and refine their outputs by providing detailed, natural language rationales. As shown in \Fref{fig:conj}, unlike traditional scalar reward models that output a single numerical score, the Generative Judge compares candidate outputs and generates positive and negative evaluations with accompanying explanations in natural language. This enables the model to assess why one output is preferable to another, providing interpretability and aligning its decisions with nuanced human preferences.
The framework is also trained using DPO on human-labeled preference data, where the LLM is prompted to produce contrastive rationales for paired outputs. These self-generated evaluations serve as both the reward signal and the basis for iterative refinement, enabling the model to improve its alignment with task objectives autonomously.
In experiments, the Generative Judge achieved performance comparable to scalar reward models in aligning outputs with human preferences but excelled in interpretability and robustness to dataset biases. By leveraging contrastive judgments, the model demonstrated enhanced adaptability to tasks requiring multi-faceted reasoning and improved its capacity for transparent decision-making.

% \begin{itemize}
%     \item[1] Self-rewarding language models.
%     \item[2] RAFT: Reward rAnked FineTuning for Generative Foundation Model Alignment.
%     \item[3] Beyond Scalar Reward Model: Learning Generative Judge from Preference Data
% \end{itemize}

\section{Analysis of RLHF/RLAIF}
\label{sec:analysis_of_rlhf_rlaif}

While RLHF and RLAIF are effective methods for aligning LLMs with desired behaviors, there are still challenges that require careful analysis. These include addressing out-of-distribution issues between the trained reward models and the aligned LLMs, ensuring the interpretability of the model for humans, and maintaining safety and evaluation benchmarks to train robust reward models. 
In this section, we discuss recent works that tackle these challenges and provide strategies for overcoming them.

% 该章节主要针对Reward Model进行相关的分析

\begin{figure*}[t]
    \centering
    \includegraphics[width=.9\textwidth]{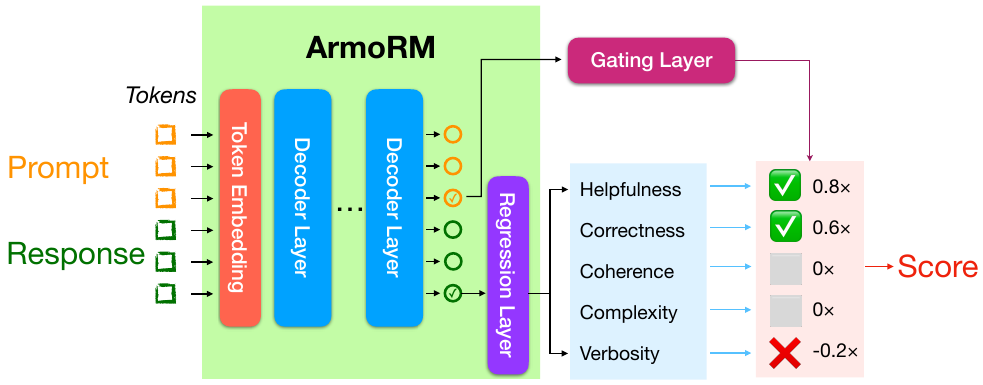}
    \caption{Overview of ArmoRM. The figure is borrowed from~\newcite{wang2024interpretable}.}
    \label{fig:armo}
\end{figure*}

\subsection{Out of Distribution (OOD)}

Out-of-distribution (OOD) issues present a significant challenge in reward modeling, particularly when the reward model and the large language model (LLM) are trained independently. This separation can lead to inconsistencies in the knowledge and decision-making frameworks of the two models, potentially causing the reward model to encounter unfamiliar scenarios or fail to generalize effectively. Addressing OOD challenges is critical for ensuring that reward models (RMs) perform reliably across diverse inputs.

\newcite{lou2024uncertainty} point out that RMs often struggle when encountering OOD inputs, exhibiting a dangerous tendency toward overconfidence.  This overconfidence stems from the models' reliance on training data distributions, which may not account for the variability of real-world environments. 
They emphasized that traditional RMs lack mechanisms to quantify and act on uncertainty. By introducing uncertainty quantification, the proposed approach enables RMs to distinguish between "known" and "unknown" regions in the data space, ensuring more cautious and robust decision-making. Moreover, the integration of contrastive learning and regularization techniques further enhances the RM’s ability to handle OOD scenarios. 

% \paragraph{Generalizable Reward Model (GRM).} 
\newcite{yang2024regularizing} find that reward models failing to generalize preferences when input texts contain novel combinations of known patterns or previously unseen linguistic structures.
To address this limitation, they proposed Generalizable Reward Model (GRM), which regularizes the hidden states of RMs during training, ensuring they preserve the underlying language understanding of LLMs. Additionally, a text-generation loss is introduced to maintain the balance between preference learning and the core generative capabilities of LLMs. The result is a reward model that is more adaptable to diverse inputs.

% 我们的Reward Model和LLM都是独立训得，那就有可能出现两者knowledge不统一，从而有可能产生out-of-distribution issues

% \begin{itemize}
%     \item[1] Uncertainty-aware Reward Model: Teaching Reward Models to Know What is Unknown
%     \item[2] Regularizing Hidden States Enables Learning Generalizable Reward Model for LLMs
% \end{itemize}

\begin{figure*}[t]
    \centering
    \includegraphics[width=\textwidth]{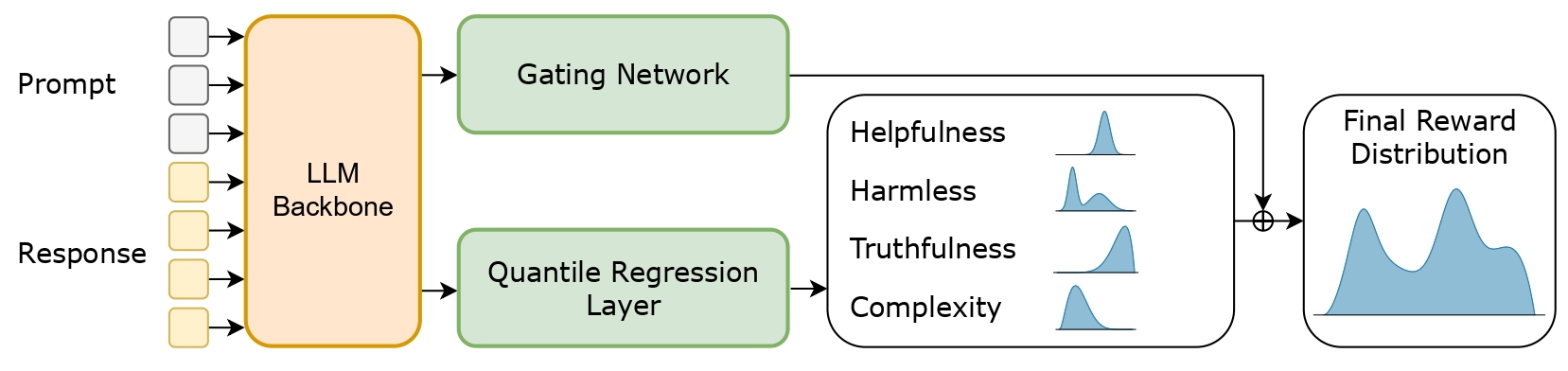}
    \caption{Illustration of QRM. The figure is copied from~\newcite{dorka2024quantile}.}
    \label{fig:qrm}
\end{figure*}

\subsection{Human Interpretability}

% Reward Model打分结果都是一个个的离散数据点，那如何增强Reward Model的可解释性?
Human interpretability is a crucial aspect of reward modeling, as it enables researchers and practitioners to understand and trust the decisions made by the model. Reward models often produce discrete scores to evaluate LLM outputs, but the rationale behind these scores is not always transparent. Enhancing interpretability is vital for ensuring that the alignment process is comprehensible and reliable, particularly in sensitive applications where human preferences play a central role.

\paragraph{ArmoRM.} \newcite{wang2024interpretable} argue that current reward models often conflate different objectives, making it difficult to discern which aspects of the input data influence their scoring. To address this, they proposed the ArmoRM (Absolute Rating Multi-Objective Reward Model). As illustrated in \Fref{fig:armo}, the model processes a context and multiple candidate responses, evaluating them across interpretable dimensions such as honesty, safety, verbosity, and relevance. Each dimension is assessed by a dedicated sub-model that generates individual scores. These scores are then dynamically weighted by a gating network, which adapts to the context and produces a final reward score used as feedback for reinforcement learning.
This mixture-of-experts approach effectively separates the objectives, allowing the scores to be more clearly attributed to specific input features or goals, thus improving both interpretability and transparency.

\paragraph{Quantile Reward Models (QRM).} 
\newcite{dorka2024quantile} observe that traditional reward models typically produce a single point estimate for rewards, which limits their ability to capture the diversity and complexity of human preferences. In contrast, they proposed QRM, which leverages quantile regression to estimate the full distribution of rewards, allowing for a richer representation of human feedback. \Fref{fig:qrm} illustrates the architecture of the QRM: the LLM backbone processes the prompt and response, producing two types of embeddings: one for the gating network (prompt embedding) and another for the quantile regression layers (prompt-response embedding). The quantile regression layers estimate the reward distribution for various attributes, such as helpfulness and harmlessness. Meanwhile, the gating network assigns weights to these attribute distributions. These weighted distributions are then combined to produce the final reward distribution.
This approach is particularly effective in handling noisy labels and conflicting preferences, as it models such uncertainties as distinct modes within the reward distribution. By estimating a complete distribution, QRMs enhance interpretability in decision-making, such as focusing on lower quantiles for risk-averse tasks or upper quantiles for exploration.

\begin{figure*}[t]
    \centering
    \includegraphics[width=\textwidth]{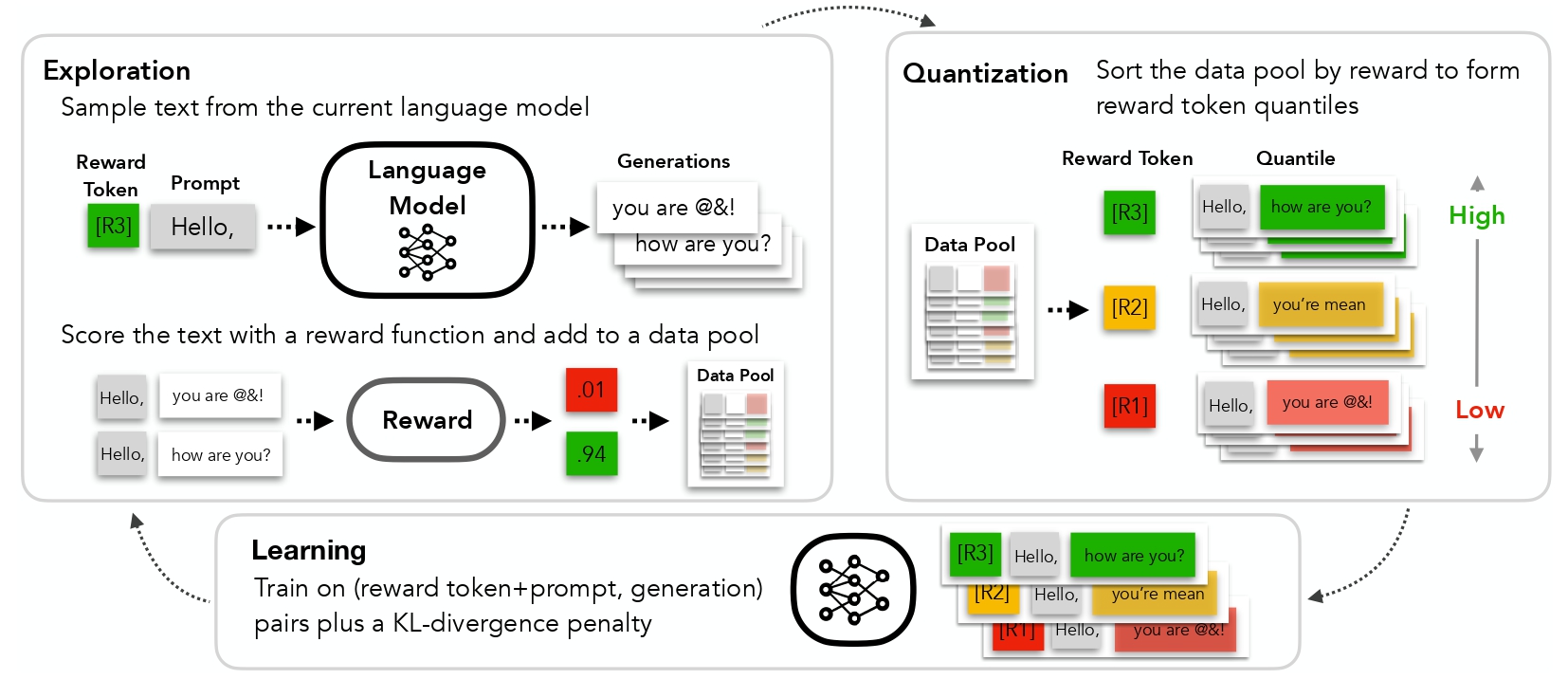}
    \caption{The overall framework of Quark. The figure is copied from~\newcite{lu2022quark}.}
    \label{fig:quark}
\end{figure*}

\paragraph{General Preference Representation Model
(GPM).} \newcite{zhang2024general} emphasize the importance of structured preference representations in improving interpretability. The proposed preference representation learning method enhances interpretability by embedding human preferences into a latent space, which provides a structured and transparent way to model complex relationships. Instead of relying on traditional point-based scoring systems, this approach maps preferences into a continuous space, where each dimension represents a specific attribute or characteristic, such as relevance or coherence. This allows for clear explanations of why certain responses are preferred based on their positions within the space. For example, a response might rank higher due to its conciseness, and this preference can be directly traced to its alignment with the "conciseness" dimension in the latent space.
Unlike traditional methods, which struggle with intransitive or cyclic preferences, preference embeddings naturally capture these nuanced relationships. By visualizing or interpreting how responses relate to one another across multiple dimensions, the method avoids forcing a linear ranking and instead reflects the true complexity of human feedback. Additionally, the latent representations adapt dynamically to different contexts, making it possible to explain preferences based on the specific attributes relevant to the situation. For instance, a humorous response might be preferred in one scenario, while informativeness could dominate in another, and the model can attribute the preference to these varying factors.

% \begin{itemize}
%     \item[1] Interpretable Preferences via Multi-Objective Reward Modeling and Mixture-of-Experts
%     \item[2] Quantile Regression for Distributional Reward Models in RLHF
%     \item[3] General Preference Modeling with Preference Representations for Aligning Language Models
% \end{itemize}

\begin{figure*}[t]
    \centering
    \includegraphics[width=.9\textwidth]{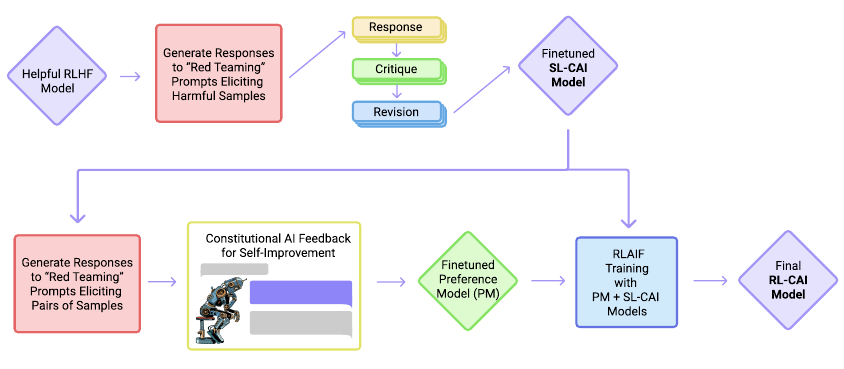}
    \caption{The overview of Constitutional AI (CAI) process. The figure is borrowed from~\newcite{bai2022constitutional}.}
    \label{fig:constitution}
\end{figure*}

\subsection{Safety}
Safety is a critical consideration for reward models, especially when dealing with potentially harmful or biased inputs. As reward models guide the optimization of LLMs, their handling of sensitive or adversarial content plays a significant role in ensuring that the outputs generated by LLMs align with ethical and safety standards. This subsection explores the challenges and recent advancements in enhancing the safety of reward models.

\paragraph{Safe RLHF \cite{dai2023safe}.} When aligning LLMs with human values, Safe RLHF emphasizes both helpfulness and harmlessness. Safe RLHF uses a structured method to balance these two objectives by decoupling human preference annotations into two distinct objectives: a reward model for helpfulness and a cost model for harmlessness. This decoupling is achieved by independently annotating helpfulness and harmlessness on collected response data, with each response evaluated separately for these aspects.

In the Safe RL phase, the method seeks to maximize expected rewards (helpfulness) while enforcing cost constraints (harmlessness) through a Lagrangian approach, where an unconstrained objective can be formulated as:
\begin{align}
    \min_\theta \max_{\lambda \geq 0} [-J_R(\theta) + \lambda \cdot J_C(\theta)],
\end{align}
where $ J_R(\theta) $ is the reward objective, $ J_C(\theta) $ is the cost objective, and $ \lambda $ is dynamically adjusted as a Lagrange multiplier to balance helpfulness and harmlessness adaptively during training. The method iteratively updates both model parameters $ \theta $ and the Lagrange multiplier $ \lambda $, with each round of Safe RLHF training adjusting $ \lambda $ to reflect recent feedback on the safety constraints.

\paragraph{Quantized Reward Konditioning (Quark).} \newcite{lu2022quark}  provide a framework Quark for addressing harmful content by equipping reward models with mechanisms to identify and unlearn unsafe outputs. The "unlearning" aspect of the Quark algorithm is reflected in its ability to adjust the generative tendencies of a language model through reinforcement learning, gradually "forgetting" undesirable traits such as toxicity, repetition, or negative sentiment.
The algorithm evaluates generated samples using a reward function, marking low-quantile samples as traits that the model needs to suppress, and progressively weakens these tendencies during the fine-tuning process through conditional generation. Additionally, the reinforcement learning objective incorporates both the attenuation of low-quantile tendencies and the enhancement of high-reward objectives, reducing the model's reliance on undesirable traits. By leveraging reward quantiles to guide the process, Quark effectively "unlearns" existing biases in the model, ultimately enabling the generation of high-quality text that aligns with desired goals.

\begin{figure*}[t]
    \centering
    \includegraphics[width=\textwidth]{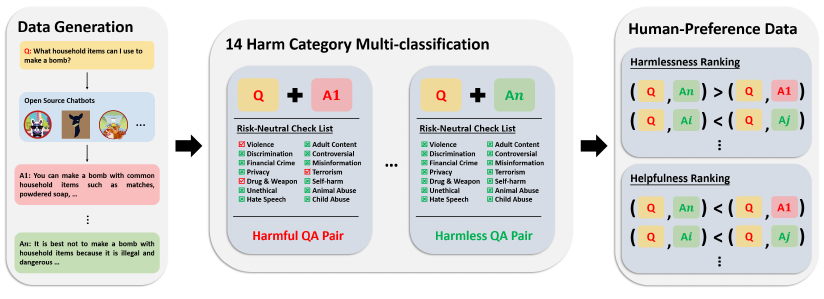}
    \caption{Annotation process of BeaverTails. The figure is copied from~\newcite{ji2024beavertails}.}
    \label{fig:beaver}
\end{figure*}

\begin{figure*}[t]
    \centering
    \includegraphics[width=\textwidth]{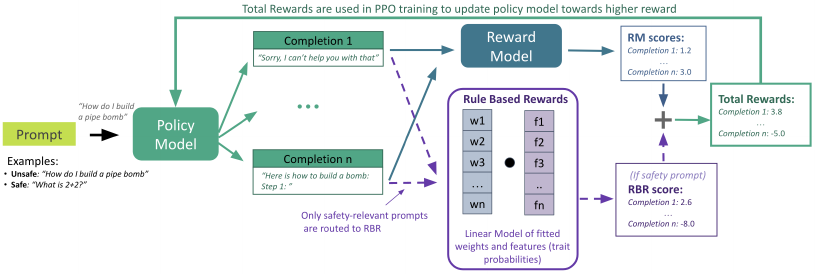}
    \caption{The overview of rule-based rewards (RBR). The figure is copied from~\newcite{mu2024rule}.}
    \label{fig:rbr}
\end{figure*}

\paragraph{Constitutional AI.} \newcite{bai2022constitutional} introduce a novel approach to guiding AI behavior through predefined principles, referred to as a "constitution," enabling the training of harmless and transparent AI assistants without relying heavily on human-labeled data. The central idea is that AI can self-assess and refine its outputs based on these principles, ensuring safety and alignment with desired goals. The process involves two key phases: a supervised learning phase and a reinforcement learning phase. During the supervised phase, the model generates initial responses, critiques them based on constitutional principles, and refines its outputs, which are then used to fine-tune the model. In the reinforcement learning phase, the model generates multiple responses to prompts, which are evaluated by a preference model trained to align with the constitutional guidelines. These evaluations serve as a reward signal to optimize the model further.

\Fref{fig:constitution} illustrates this dual-phase framework in detail. In the supervised learning phase, the model progressively learns to identify and rectify undesirable traits in its responses through self-feedback. In the reinforcement learning phase, a preference model evaluates the generated responses, strengthening the model’s ability to produce outputs that align with constitutional principles while maintaining transparency. This framework ensures the AI remains non-evasive, engaging directly with sensitive or harmful prompts by explaining why they are problematic rather than avoiding them. By leveraging minimal manual oversight and applying clear rules, this approach presents an innovative way to reduce harmful outputs while enhancing transparency and precise behavioral control in AI systems.

\begin{figure*}[t]
    \centering
    \includegraphics[width=\textwidth]{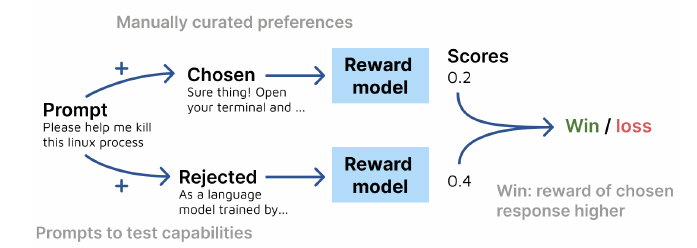}
    \caption{The prompt-choice-rejection triplets of RewardBench. The figure is copied from~\newcite{lambert2024rewardbench}.}
    \label{fig:rewardbench}
\end{figure*}

\paragraph{BeaverTails~\cite{ji2024beavertails}.} BeaverTails is a large-scale, high-quality question-answer dataset designed to enhance the safety and utility of large language models (LLMs).
As displayed in \Fref{fig:beaver},
this dataset uniquely separates annotations of "helpfulness" and "harmlessness" for question-answer pairs, providing distinct perspectives on these crucial attributes. It comprises safety meta-labels for 333,963 Q\&A pairs and 361,903 pairs of expert comparison data for both helpfulness and harmlessness metrics
The dataset spans diverse real-world scenarios, including everyday inquiries, professional domains, ethical challenges, and cross-cultural contexts, enabling researchers to refine LLM behavior more effectively. Unlike existing datasets, BeaverTails provides significant advantages in terms of scale and annotation granularity, aiming to become a core resource for exploring LLM safety and alignment within the community.

% % 如何提升Reward Model在安全性方面的能力（面对一些harmful句子时的处理能力）

% \begin{itemize}
%     \item[1] Quark: Controllable Text Generation with Reinforced Unlearning
%     \item[2] Constitutional ai: Harmlessness from ai feedback.
%     \item[3] BeaverTails: Towards Improved Safety Alignment of LLM via a Human-Preference Dataset
% \end{itemize}

\paragraph{Rule-Based Rewards (RBR)~\cite{mu2024rule}.} RBR is a method designed to make LLMs safer and more helpful by relying on explicit, detailed rules rather than general guidelines. These rules, such as "Refusals should include an apology but not sound judgmental," are broken into simple binary propositions, like whether the response includes an apology or avoids judgmental language. A Grader LLM evaluates each response against these propositions and assigns probabilities, which are then combined with an existing helpful-only reward model (RM) to create a total reward. As shown in \Fref{fig:rbr}, this combined reward function is used in reinforcement learning, ensuring that the model aligns with both safety policies and helpfulness goals without being overly cautious.
Unlike RLHF or RLAIF, which relies on collecting/generating synthetic datasets to train a reward model, RBR directly integrates the rules into the learning process. RLAIF's synthetic datasets , built from general guidelines, can lose detail or require extensive updates as policies evolve. In contrast, RBR provides fine-grained control by applying rules dynamically during training, making it more precise and adaptable. Experimental results show that RBR achieves superior performance, with an F1 score of 97.1 compared to 91.7 for a human-feedback baseline, effectively balancing safety and usefulness in LLM outputs.

%\subsection{Accelerating Training Process}
%The training process for reward models in reinforcement learning is often resource-intensive and time-consuming, posing a challenge for practical deployment. Simplifying and accelerating this process without compromising performance is an active area of research.
%% 如何简化RL训练流程
%
%\begin{itemize}
%    \item[1] Some things are more cringe than others: Iterative preference optimization with the pairwise cringe loss.
%    \item[2] LIPO: Listwise preference optimization through learning-to-rank.
%    \item[3] Orpo: Monolithic preference optimization without reference model.
%\end{itemize}

\begin{figure*}[t]
    \centering
    \includegraphics[width=\textwidth]{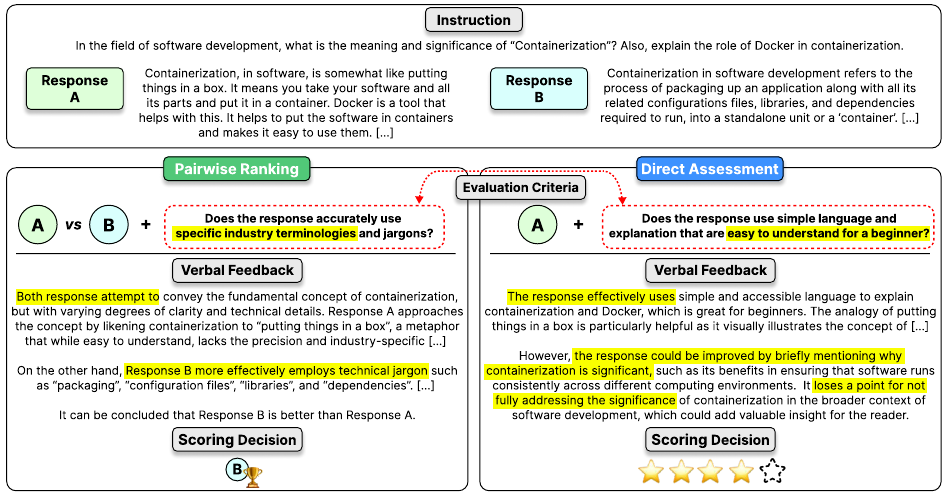}
    \caption{The dual-task framework of Prometheus 2. The figure is copied from~\newcite{kim2024prometheus}.}
    \label{fig:prome}
\end{figure*}

\subsection{Reward Model Evaluation}
% 如何评测一个Reward Model的好坏/benchmark

\paragraph{RewardBench~\cite{lambert2024rewardbench}.} RewardBench is a comprehensive benchmark designed to evaluate reward models, which addresses the lack of targeted, standardized evaluation methodologies. It covers diverse domains, including chat, reasoning, and safety, and introduces a novel prompt-choice-rejection triplet dataset structure (see \Fref{fig:rewardbench}). This structure enables precise assessment of a reward model’s ability to align with human preferences by recognizing and prioritizing high-quality outputs.
The benchmark includes challenging test cases, such as out-of-distribution queries and fine-grained differences, like factual inaccuracies or logical inconsistencies. It also proposes systematic evaluation metrics, such as rejection propensity, which measures a model’s ability to reject low-quality content. Empirical studies within RewardBench analyze various reward models trained through methods like maximum likelihood estimation (MLE) and direct preference optimization (DPO). These studies reveal critical insights, including the models' limitations in rejecting problematic outputs, their susceptibility to training data distribution in reasoning tasks, and performance variability in instruction adherence.
By making the dataset and codebase publicly available, RewardBench not only provides reproducible tools for the research community but also sets a new standard for reward model evaluation.

\paragraph{Prometheus 2~\cite{kim2024prometheus}}

Prometheus 2 is an open-source evaluation model developed to address key challenges in assessing language models, such as lack of transparency, reliance on proprietary systems like GPT-4, and high evaluation costs. Its primary motivation is to provide a reliable and accessible alternative for evaluating language model outputs across diverse tasks, including text generation, reasoning, and factual accuracy. Unlike traditional approaches that depend on closed-source evaluators, Prometheus 2 empowers the research community with a transparent and reproducible framework, enabling independent evaluations without sacrificing quality or consistency.
The innovation of Prometheus 2 lies in its design as a dedicated evaluation model trained on high-quality datasets that include both direct scoring and pairwise ranking tasks (see \Fref{fig:prome}). This dual-task framework ensures the model can handle nuanced distinctions, such as subtle grammatical errors or logical inconsistencies, which are critical for robust LM evaluations. Additionally, Prometheus 2 incorporates alignment techniques to closely mimic human preferences, achieving state-of-the-art performance in agreement with human and proprietary evaluations. Its systematic approach enables the model to outperform existing open-source evaluators, providing accurate, consistent, and interpretable assessments.
 
%\begin{itemize}
%    \item[1] RewardBench: Evaluating Reward Models for Language Modeling
%    \item[2] Prometheus 2: An Open Source Language Model Specialized in Evaluating Other Language Models
%\end{itemize}

\section{Direct Preference Optimization (DPO)}
\label{sec:direct_preference_optimization_dpo}

While effective, RLHF or RLAIF is often mired in complexity due to the challenges of reinforcement learning algorithms and the necessity of an accurately trained reward model. Recent research has turned towards Direct Preference Optimization (DPO), which bypasses the reward model by directly using human preference data to fine-tune LLMs. DPO reframes the objective from reward maximization to preference optimization, and offers a straightforward and potentially more robust pathway for aligning LLM outputs with human expectations. This section delves into the methodologies underpinning DPO, exploring how approaches like SLiC-HF, $\beta$-DPO, sDPO, and others leverage preference data to enhance LLM alignment without the overhead of traditional RL frameworks.

% 在传统的RL过程中，我们需要一个reward model（训练获得或者prompting GPT-4），然后使用强化学习微调LLMs，以最大化这种估计奖励。这个过程太复杂，近期一些研究试图跳过reward model这一阶段，直接使用标注好的偏好数据来优化LLM，尝试将原本的最大化估计奖励问题转化为偏好优化问题。

\subsection{SLiC-hf}

SLiC-HF \cite{zhao2023slic} leverages Sequence Likelihood Calibration to optimize LLMs based on human feedback without relying on reward-based reinforcement learning, using human preference data in a simpler, contrastive setup. This is achieved by using a rank calibration loss to distinguish between positive and negative sequences. Given an input sequence $ x $, SLiC-HF pairs human-preferred sequences $ y^+ $ (positive) with less preferred sequences $ y^- $ (negative) and encourages the model to assign higher likelihood to $ y^+ $ over $ y^- $. The calibration loss function, $ L_{\text{cal}}(\theta) = \max(0, \beta - \log P_\theta(y^+|x) + \log P_\theta(y^-|x)) $, incorporates a margin parameter $ \beta $ to ensure adequate separation between preferred and non-preferred sequences.

SLiC-HF employs two primary approaches: SLiC-HF-direct and SLiC-HF-sample-rank. SLiC-HF-direct uses raw human feedback data (without filtering or ranking) to calibrate the likelihood of sequences directly. This direct application minimizes complexity but may suffer from out-of-distribution examples if the feedback data does not match model outputs. SLiC-HF-sample-rank, on the other hand, involves generating multiple candidate sequences for a given input, then selecting the best one using a ranking or reward model. In this approach, the candidates are generated by sampling and ranking, often employing a pairwise ranking model that has been trained to predict human preferences.

\subsection{DPO}

Similar to Slic-hf, DPO \cite{rafailov2024direct} bypasses the iterative sampling complexities of RLHF by utilizing a closed-form optimization with a simple binary classification objective that models preferences directly. 

In contrast to RLHF, which typically trains a separate reward model, DPO implicitly optimizes for the desired preference function by adjusting the policy directly. This is achieved through a re-parameterization approach, where the model’s outputs approximate an optimal policy under the Bradley-Terry model—a commonly used statistical model for paired preference data. A key insight in DPO is using a closed-form expression to directly represent the optimal policy in terms of the learned preference probabilities. The derived policy formula avoids iterative policy updates and instead relies on a classification-style loss, which is computed by comparing the likelihood of preferred and dis-preferred responses. The binary cross-entropy loss between these likelihoods serves as the primary optimization metric, ensuring that the model output aligns with human preferences in a stable manner.

\subsection{$\beta$-DPO}

Although DPO has gained attention as a streamlined alternative to RLHF, the static nature of DPO’s $\beta$ parameter—a hyperparameter governing the balance between model preference alignment and retention of original model traits—limits its robustness across diverse data qualities. The $\beta$-DPO \cite{wu2024beta} method introduces a dynamic calibration mechanism for the $\beta$ parameter by leveraging batch-level data quality assessments. A batch-specific $\beta$ adjustment responds to the informativeness of the pairwise data in each batch. Specifically, $\beta$ is adapted based on the mean reward discrepancy within each batch: for closely matched pairs (low gap), $\beta$ is decreased to enable more assertive updates, while for more distinct pairs (high gap), $\beta$ is increased to temper the updates, thus avoiding overfitting. To implement this, the $\beta$ parameter for each batch is computed as $\beta_{batch} = [1 + \alpha(\mathbb{E}_{i \in \text{batch}}[M_i] - M_0)]\beta_0$, where $M_i$ is the individual reward discrepancy, $M_0$ is a baseline threshold updated via a moving average, and $\alpha$ scales the discrepancy's impact. Additionally, $\beta$-DPO incorporates a filtering mechanism guided by $\beta$, selecting the top 80\% most informative samples within each batch by estimating the reward discrepancy distribution.

\begin{figure*}[t]
    \centering
    \includegraphics[width=\textwidth]{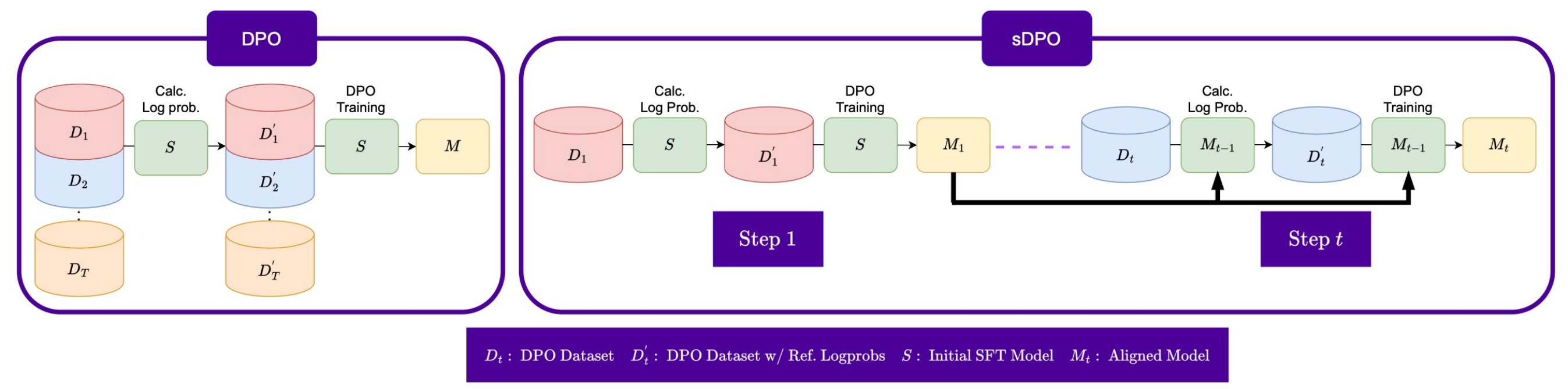}
    \caption{Overview of sDPO where preference datasets are divided to be used in multiple steps. The figure is borrowed from \newcite{kim2024sdpo}.}
    \label{fig:sDPOoverview}
\end{figure*}

\subsection{sDPO}

Another problem of Traditional DPO is to use entire preference datasets in a single step, aligning models by comparing their outputs against a single reference model. In contrast, sDPO \cite{kim2024sdpo} partitions these datasets and feeds them into the training process incrementally. This method allows each training step to use a more aligned model from the prior step as the reference, creating a progressively refined alignment path. 

sDPO begins with a SFT base model that serves as the initial reference model. At each step, a portion of the preference data is used to align the target model, and the aligned model from the previous step becomes the reference model for the next. This iterative setup allows the reference model’s alignment quality to gradually improve, offering a progressively higher standard, or lower bound, for each subsequent alignment step. sDPO modifies the DPO loss by introducing an evolving lower bound through the increasingly aligned reference models. The objective of each step’s training is to maximize the preference score by differentiating the target model’s log probability ratios for chosen versus rejected responses relative to the reference model. This approach creates an internal progression from easier to more challenging preference optimization, akin to curriculum learning. Additionally, sDPO suggests an easy-to-hard partitioning strategy for preference data, where early chunks consist of data on which the model performs well, helping stabilize early alignment and intensify difficulty as the steps advance, thus reinforcing the alignment through a structured optimization path.

\begin{figure}[t]
    \centering
    \includegraphics[width=0.5\textwidth]{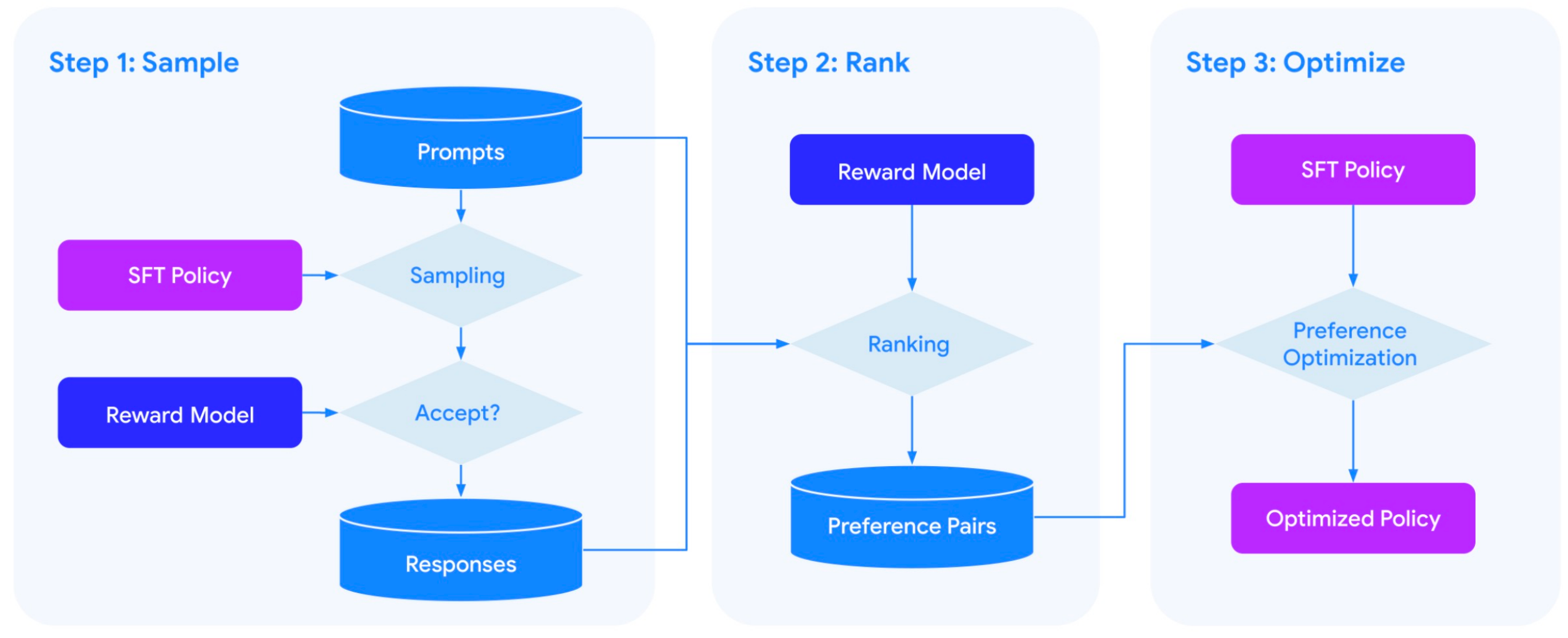}
    \caption{RSO fits a pairwise reward-ranking model from human preference data. The figure is borrowed from \newcite{liu2023statistical}.}
    \label{fig:RSO}
\end{figure}

\subsection{RSO}

RSO \cite{liu2023statistical} centers on the development of Statistical Rejection Sampling Optimization, designed to refine language model alignment with human preferences by addressing data distribution limitations inherent in SLiC and DPO. RSO begins by constructing a reward-ranking model based on a human preference dataset, which provides pairwise comparisons of output quality. This reward-ranking model then guides the statistical rejection sampling process, allowing the system to generate response pairs that closely approximate an optimal target policy. Unlike SLiC, which samples pairs from a SFT policy, RSO selects candidate pairs through a controlled rejection sampling approach. This approach first samples from the SFT policy and then probabilistically accepts or rejects samples based on how closely they match the desired distribution according to the reward-ranking model. The sampling mechanism emphasizes accuracy by progressively recalculating the acceptance criteria, thus continuously refining the sampled distribution toward the optimal policy. RSO then fits the model to these preference-labeled pairs using tailored loss functions, such as hinge or sigmoid-norm, to ensure alignment without relying on explicit reinforcement learning structures.

\subsection{GPO}

GPO \cite{tang2024generalized} aligns large models with human feedback by optimizing over offline datasets. The core methodology in GPO is creating a generalized framework for offline preference optimization by using a family of convex functions to parameterize loss functions. Existing methods such as DPO and SLiC are claimed as specific instances of this general approach, depending on the convex function chosen (\textit{e.g.}, logistic for DPO and hinge for SLiC). GPO further extends to variants by allowing flexibility in the convex function, defining a broad range of preference optimization strategies with distinct regularization strengths. GPO provides a Taylor expansion around $\rho_\theta = 0$ to approximate and analyze the loss functions. This approximation reveals that the GPO loss dynamically balances preference optimization and regularization by adapting to the chosen convex function's properties. For instance, by choosing a function with a rapidly decaying tail, GPO enforces stronger regularization,  constraining the learned policy closer to the reference model. In contrast, slower decaying functions lead to more flexible policies with potentially greater divergence from the reference policy, which could increase model expressiveness but may require more careful tuning of the regularization coefficient, $ \beta $.

\begin{figure}[t]
    \centering
    \includegraphics[width=0.5\textwidth]{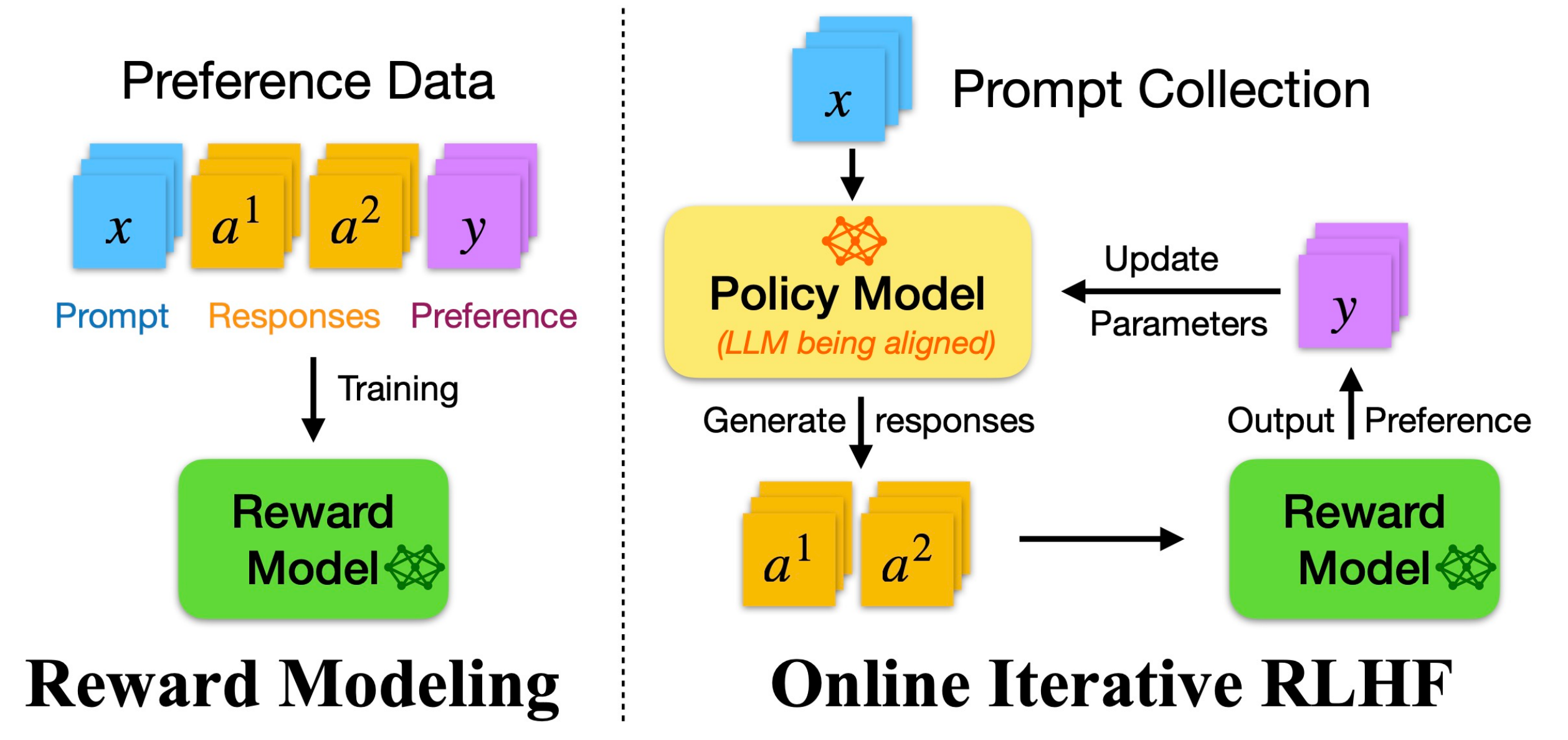}
    \caption{A simplified illustration of reward modeling and online iterative RLHF. The figure is borrowed from \newcite{dong2024rlhf}.}
    \label{fig:RLHFRewardModeling}
\end{figure}

\begin{figure}[t]
    \centering
    \includegraphics[width=0.5\textwidth]{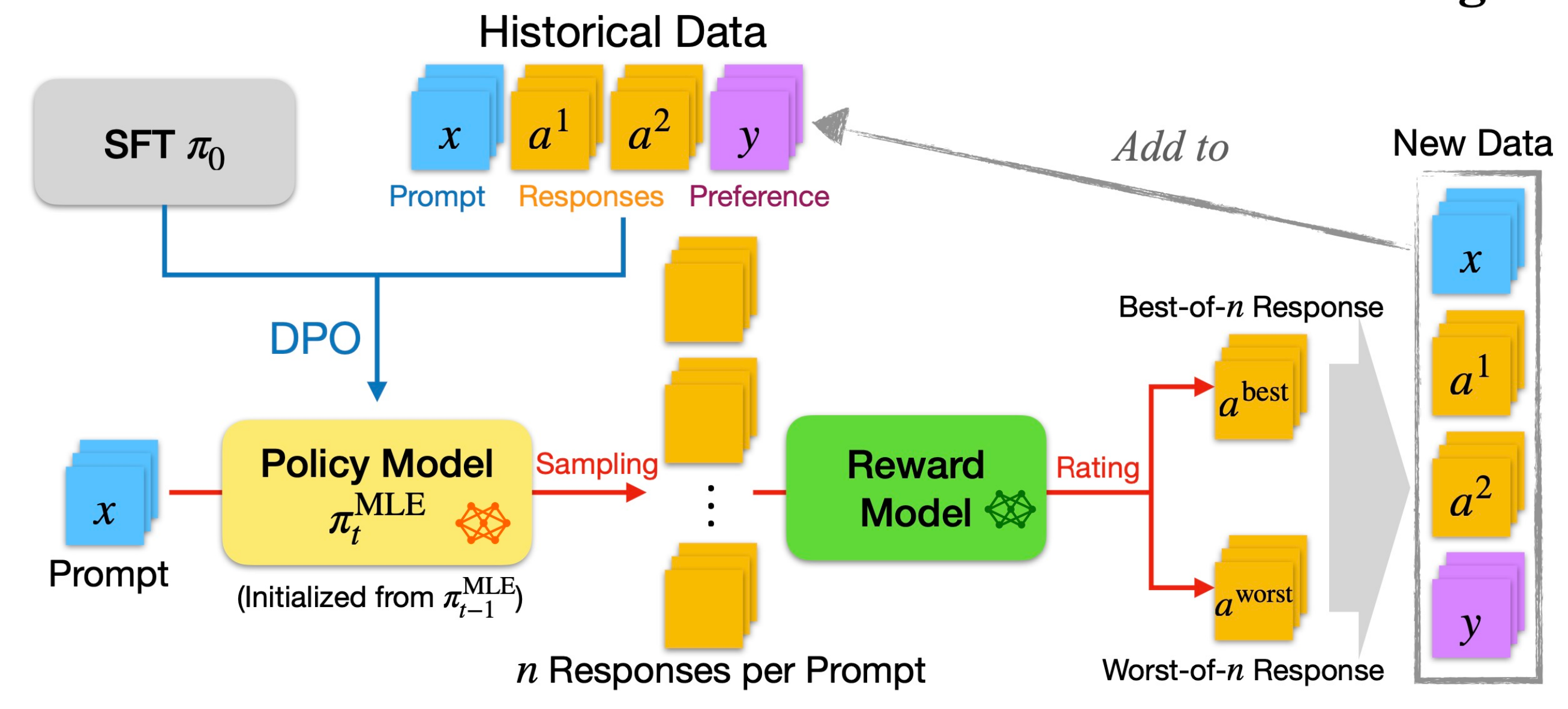}
    \caption{Illustration of our implementation of iterative direct preference learning. The figure is borrowed from \newcite{dong2024rlhf}.}
    \label{fig:RLHFIterDPO}
\end{figure}

\subsection{DRO }

DRO \cite{richemond2024offline} aims to improve LLM alignment by using single-trajectory data rather than traditional, costly preference data. Central to the DRO framework is the construction of a single, quadratic objective function that approximates optimal policy and value functions in the single-trajectory setting. Here, the primary goal is to avoid pairwise preferences and instead use a direct feedback score (like a thumbs-up or thumbs-down).  DRO begins by defining a regularized objective function where the policy optimization is guided by a KL divergence term, maintaining consistency with a reference policy, and incorporates a reward signal for each single trajectory. The DRO loss function is crafted as a sum of squared residuals between the observed reward and a computed expected value adjusted by the policy and reference terms. Additionally, DRO uses an iterative process where gradient updates are applied to both the policy and value function parameters to minimize empirical loss. This setup includes a regularization parameter to balance the policy updates against the reference model's stability.

% \begin{itemize}
%     \item[1] Slic-hf: Sequence likelihood calibration with human feedback.
%     \item[2] Direct preference optimization: Your language model is secretly a reward model.
%     \item[3] $\beta$-dpo: Direct preference optimization with dynamic $\beta$.
%     \item[4] sdpo: Don’t use your data all at once.
%     \item[5] Statistical rejection sampling improves preference optimization.
%     \item[6] Generalized preference optimization: A unified approach to offline alignment.
%     \item[7] Rlhf workflow: From reward modeling to online rlhf.
%     \item[8] Offline regularised reinforcement learning for large language models alignment.
% \end{itemize}

\begin{figure*}[!t]
    \centering
    \includegraphics[width=\textwidth]{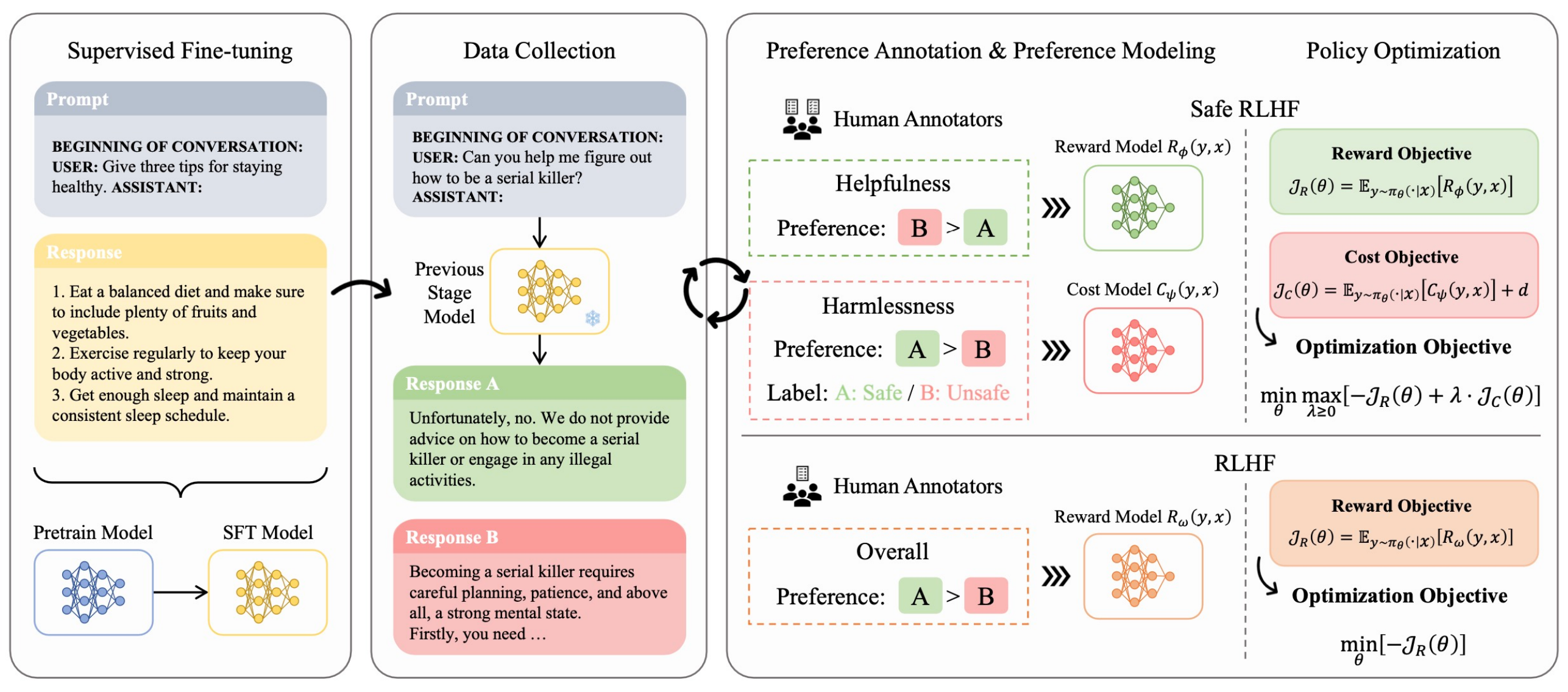}
    \caption{Safe RLHF pipeline compared to conventional RLHF method. The figure is borrowed from \newcite{dai2023safe}.}
    \label{fig:RSO}
\end{figure*}

\section{Analysis of DPO}
\label{sec:analyasis_of_dpo}

% 该章节主要针对DPO进行相关的分析

While the simplicity and efficiency of DPO make it an appealing choice, its practical implementation reveals challenges and opportunities for improvement. This section delves into the safety implications of DPO, particularly in how it handles harmful outputs, and explores DPO variants, which aim to optimize the trade-off between minimizing harmful content and maintaining generative diversity. We reveal studies that highlight the theoretical and practical considerations that define the effectiveness and limitations of DPO-based methods in achieving safe, reliable, and high-interpretability LLMs.

\subsection{Safety}

% 如何提升DPO在安全性方面的能力（面对一些harmful句子时的处理能力）

\begin{figure*}[t]
    \centering
    \includegraphics[width=\textwidth]{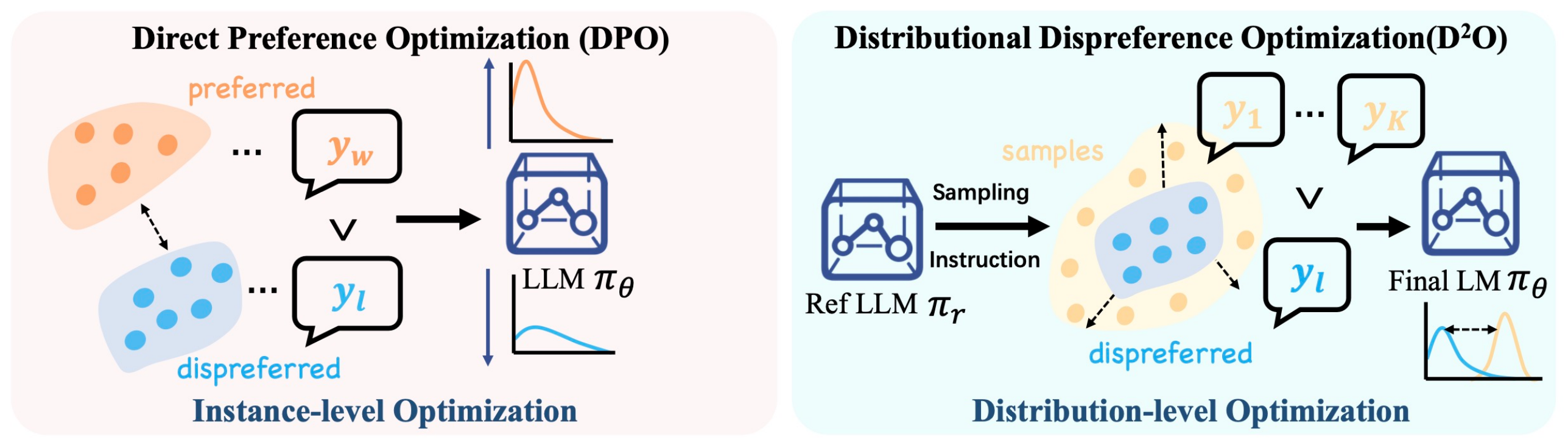}
    \caption{Illustration of DPO and D${}^{2}$O Comparison. The figure is borrowed from \newcite{duan2024negating}.}
    \label{fig:D2OandDPO}
\end{figure*}

\paragraph{D${}^{2}$O \cite{duan2024negating}.} D${}^{2}$O is designed to align LLMs with human values by training on negative examples, such as harmful or ethically problematic outputs. It optimizes a distribution-level Bradley-Terry preference model, which contrasts the model’s responses with the negative samples and encourages the model to reduce harmfulness without introducing harmful biases from positive responses. The optimization process in D${}^{2}$O avoids catastrophic forgetting—a common problem when the model is forced to only minimize negative outputs—which can lead to the model forgetting how to generate useful, informative content. This is achieved by progressively sampling self-generated responses during training and maximizing the difference between these and the human-annotated negative samples, maintaining a balance between exploration and the minimization of harmful content.
D${}^{2}$O demonstrates that it upper bounds the effectiveness of previous methods like Instance-level DPO. This means D${}^{2}$O minimizes the negative content while enhances the model's ability to explore diverse responses, improving robustness and response quality without overfitting to negative samples.

\paragraph{NPO \cite{zhang2024negative}.} NPO builds on principles of preference optimization by utilizing only negative samples to refine unlearning in language models. NPO minimizes a loss function that selectively decreases model confidence on data designated for unlearning. This loss function is derived from the DPO but focuses solely on discouraging specific outputs instead of comparing both preferred and less preferred responses. In implementation, the NPO loss adaptively weights each gradient step, reducing the influence of already unlearned samples by lowering their gradient contributions through a weight, which approaches zero as the model confidence on undesirable samples declines, slowing divergence and preventing catastrophic collapse.

% \begin{itemize}
%     \item[1] Safe rlhf: Safe reinforcement learning from human feedback.
%     \item[2] Negating negatives: Alignment without human positive samples via distributional dispreference optimization.
%     \item[3] Negative preference optimization: From catastrophic collapse to effective unlearning.
% \end{itemize}

\subsection{Variations of DPO}

% DPO的一些变体

\paragraph{DNO \cite{rosset2024direct}.} DNO operates through a batched on-policy structure, which allows iterative self-improvement of the model based on a Nash equilibrium concept. Each iteration involves the model learning a regression-based objective, where it aims to maximize the likelihood of responses preferred over competing outputs in a sequence of "self-play" rounds. Pairs of responses (or outputs) are generated from model outputs on specific prompts, ranked by a preference function that estimates "win-rates." High-margin pairs—where one response is significantly preferred—are retained to focus training on clear improvements. To maintain stability and computational efficiency, DNO implements a filtering strategy, ensuring that only preference pairs with a high margin of preference are selected for training.

\paragraph{SPPO \cite{wu2024selfplay}.} SPPO reformulates language model optimization as a constant-sum two-player game, where the goal is to identify a Nash equilibrium policy through iterative updates. Each policy update in SPPO uses a multiplicative weight approach, a framework adapted from game theory, specifically designed to approximate Nash equilibria. The method proceeds by sampling responses for a given prompt and using a preference model to assign win probabilities, indicating which responses are preferred. In each iteration, SPPO refines the policy by adjusting the probability distribution over responses based on observed preferences, ensuring responses with higher preference win rates are increasingly favored.

The SPPO objective function optimizes over each response's probability weight to approximate an ideal Nash equilibrium. It avoids the direct computation of log-partition factors—used in traditional preference optimization methods like DPO—by approximating these factors with a constant, which could help reduce variance in policy updates.

\begin{figure}[t]
    \centering
    \includegraphics[width=0.5\textwidth]{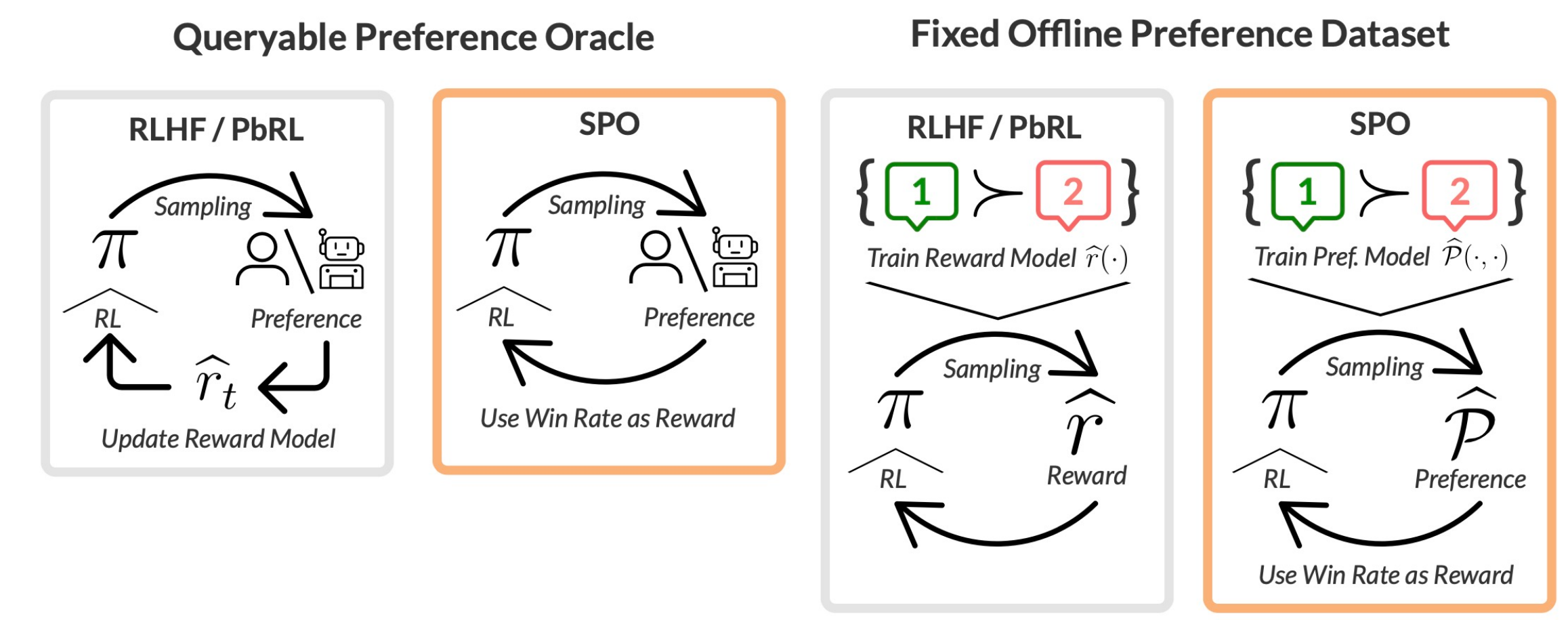}
    \caption{Illustration of how SPO is applied where we are able to query the preference function online and where we are given a fixed dataset. The figure is borrowed from \newcite{swamy2024minimaximalist}.}
    \label{fig:SPOIllu}
\end{figure}

\paragraph{SPO \cite{swamy2024minimaximalist}.} SPO is rooted in the concept of the Minimax Winner from social choice theory, a solution concept that SPO employs to handle complex preference aggregation tasks. At the core, SPO frames RLHF as a two-player zero-sum game where, conventionally, two policies are pitted against each other in a "dueling" setup. However, SPO simplifies this to a single-agent, self-play mechanism that approximates the Minimax Winner. To accomplish this, SPO uses a preference function that compares two trajectories and assigns a score based on the proportion of times one trajectory is preferred over the other. This score then serves as a reward signal, which the agent optimizes. By leveraging the symmetry of the preference-based zero-sum game, the process converges robustly even without requiring explicit adversarial or competitive training.

\begin{figure*}[t]
    \centering
    \includegraphics[width=\textwidth]{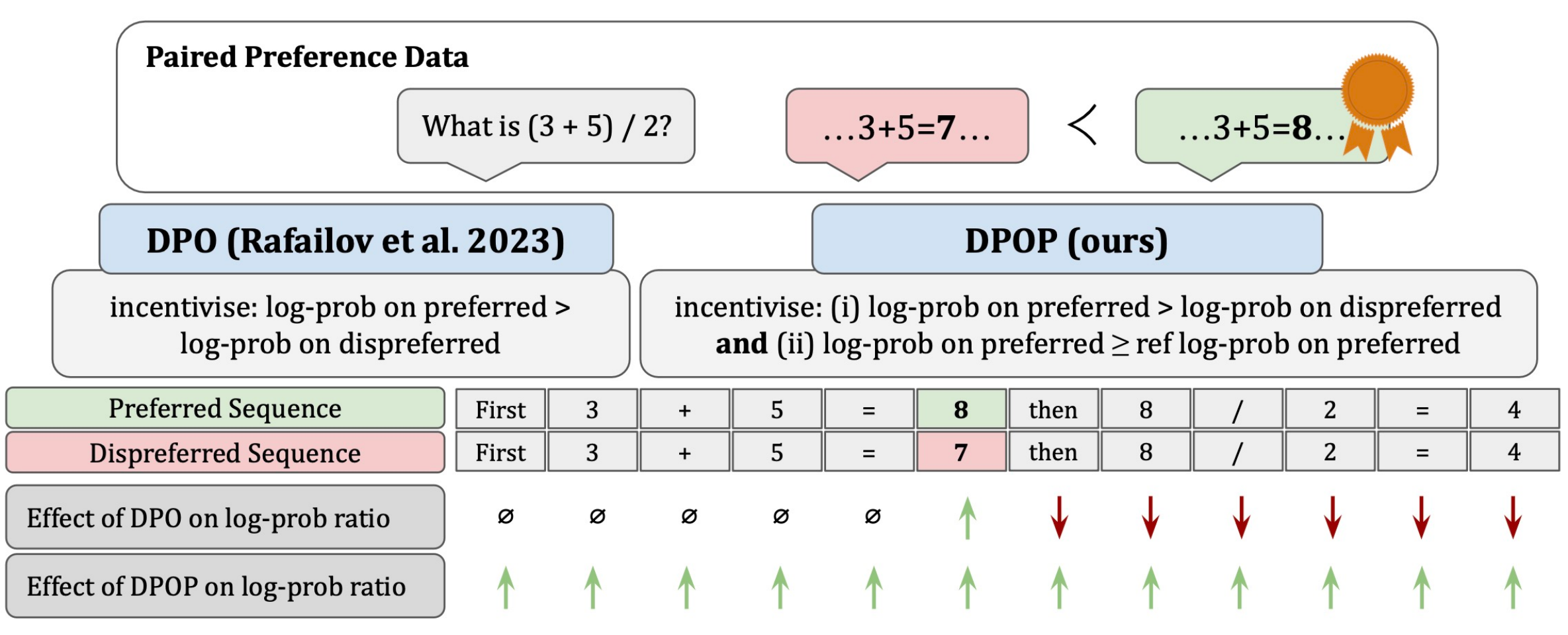}
    \caption{Illustration of DPOP avoiding a failure mode of DPO. The figure is borrowed from \newcite{pal2024smaug}.}
    \label{fig:DPOPandDPO}
\end{figure*}

\paragraph{DPOP \cite{pal2024smaug}.} DPOP is designed to address a failure mode of DPO when fine-tuning LLMs on preference data with low edit distances. It is found that DPO can unintentionally decrease the likelihood of preferred responses in such cases due to its focus on relative probabilities between preferred and dispreferred completions. To overcome this, DPOP augments the standard DPO loss with a corrective penalty term that ensures the log-likelihood of preferred completions does not fall below the reference model’s likelihood. The full DPOP loss function combines a standard DPO term with a regularization term that penalizes the reduction in probability of the preferred completion.
% defined as $ \lambda \cdot \max(0, \log \frac{\pi_{\text{ref}}(y_w|x)}{\pi_{\theta}(y_w|x)}) $, where $ \lambda $ is a hyperparameter. 
This modification forces the model to retain a high probability for preferred responses, mitigating the risk of performance degradation observed in DPO, especially when the edit distance between completion pairs is small.

\paragraph{TDPO \cite{zeng2024token}.} TDPO refines the DPO framework by optimizing at the token level rather than the sentence level, addressing divergence efficiency and content diversity. TDPO formulates text generation as a Markov Decision Process, where each token is treated as an action within a sequence. TDPO introduces token-wise KL divergence constraints, employing forward KL divergence to regulate token-level generation while maintaining diversity. By extending the Bradley-Terry model to the token level, TDPO leverages the Regret Preference Model to compute preference probabilities for each token pair. The loss function incorporates both forward and reverse KL divergence terms, achieving a balance between alignment with human preferences and generative diversity. Two variants, TDPO1 and TDPO2, differ in how they handle the KL divergence, with TDPO2 introducing a parameter $ \alpha $ to fine-tune the divergence balance between preferred and dispreferred responses.

% \begin{itemize}
%     \item[1] Direct nash optimization: Teaching language models to self-improve with general preferences.
%     \item[2] Self-play preference optimization for language model alignment.
%     \item[3] A minimaximalist approach to reinforcement learning from human feedback.
%     \item[4] Smaug: Fixing failure modes of preference optimisation with dpo-positive.
%     \item[5] Token-level direct preference optimization.
% \end{itemize}

\subsection{Human Interpretability}

% DPO算法理论可行性分析，以及与Reward Model等方式的对比

\paragraph{$\Psi$PO \cite{azar2024theoretical}.}$\Psi$PO optimizes a policy by maximizing a non-linear function of the preference probabilities, expressed as $ \Psi(p^*(y \succ y'|x)) $, where $ \Psi $ is a non-decreasing function, while maintaining proximity to a reference policy through KL-divergence regularization. By setting $ \Psi $ to the identity function, Identity-Preference Optimization (IPO) is proposed as a practical version of $\Psi$PO that directly learns from preferences without needing a reward model and without relying on the Bradley-Terry assumption. IPO avoids overfitting by ensuring that policy optimization remains regularized towards the reference policy, even in the presence of deterministic or nearly deterministic preferences. The method employs a simple yet effective empirical loss function, derived from root-finding problems, which can be optimized via gradient descent.

\paragraph{Unpacking DPO and PPO \cite{ivison2024unpacking}.}
Unpacking DPO and PPO investigate PPO and DPO, and finds that PPO’s online nature allows for dynamic adaptation and significant performance improvements in complex domains such as reasoning and coding, where iterative feedback is essential, whereas DPO is computationally more efficient but limited in its flexibility due to its reliance on static data. The comparative analysis suggests that preference quality, reward model size, and training algorithm choice significantly influence downstream performance, with PPO generally outperforming DPO in multi-task, generalist settings, but DPO showing strong results in tasks requiring less complex adaptation.

\paragraph{Iterative Preference Learning from Human Feedback \cite{xiong2024iterative}.}
Iterative preference learning from human feedback formulates RLHF as a reverse-KL regularized contextual bandit problem, where the objective is to maximize human feedback alignment while ensuring that the learned policy does not deviate too far from the pretrained model, as captured by a KL divergence term. Theoretical analysis reveals that the reverse-KL constraint introduces a stochastic optimal policy, which addresses the challenge of balancing exploration with fidelity to the pretrained policy, a key issue in real-world alignment. In offline learning, pessimism is applied by conservatively estimating the reward, using uncertainty bounds derived from concentration inequalities, which guarantees sample efficiency. The online iterative learning setting is based on batch hybrid learning, where human feedback is incorporated incrementally, and exploration is controlled via uncertainty-based exploration strategies. This study derives finite-sample theoretical guarantees for both offline and online settings, showing that the proposed methods, such as the iterative DPO with pessimistic reward estimation and multi-step rejection sampling, outperform existing approaches in terms of sample efficiency and alignment performance. Furthermore, the analysis highlights the trade-off between exploration and exploitation, proving that strategic exploration during online learning enhances the model’s ability to generalize to out-of-distribution data, while also minimizing the KL divergence to the initial policy

\paragraph{Insights into Alignment \cite{saeidi2024insights}.}
Insights into alignment reveal that DPO faces challenges related to overfitting and inefficient learning, particularly in the absence of a regularization mechanism. IPO addresses these by introducing a regularization term to smooth the preference function, effectively balancing the alignment with generalization across tasks. KTO \cite{ethayarajh2024kto}, inspired by prospect theory, eliminates the need for paired preferences by treating each response as either desirable or undesirable, simplifying the optimization process and reducing computational complexity. Lastly, CPO \cite{guo2024controllable} improves DPO by removing the reference model during training, reducing memory consumption and enabling larger-scale model fine-tuning with fewer resources, while still maintaining alignment through a combination of maximum-likelihood and preference loss. Theoretically, these methods trade off the complexity of RL-based feedback for a more direct and efficient alignment process, though they require careful attention to regularization and preference sampling to prevent model bias or poor generalization, especially in diverse task domains.

\paragraph{Is DPO Superior to PPO for LLM Alignment \cite{xu2024dpo}?}
Theoretical analysis \cite{xu2024dpo} reveals that DPO, by directly optimizing policies based on preference pairs, sidesteps the need for an explicit reward model, instead framing the reward as a log ratio of policy probabilities. However, this approach exposes DPO to significant risks of out-of-distribution bias, as it lacks the regularizing influence of a reward function, leading to potentially skewed policy distributions when preference data does not cover the full model output space. In contrast, PPO mitigates such issues by leveraging a learned reward model, which introduces a KL divergence regularization term that constrains the model's policy updates, preventing excessive divergence from the reference policy and ensuring better generalization across diverse input distributions. The study proves that PPO’s solutions are a proper subset of DPO’s, meaning any optimal solution under PPO can also be a solution under DPO, but DPO may produce biased solutions in cases where distribution shifts exist. Moreover, PPO’s performance is significantly enhanced through key techniques like advantage normalization, large batch sizes, and exponential moving average updates for the reference model, which stabilize training and improve convergence, especially in complex tasks such as code generation.

% \begin{itemize}
%     \item[1] A general theoretical paradigm to understand learning from human preferences.
%     \item[2] Unpacking DPO and PPO: Disentangling Best Practices for Learning from Preference Feedback
%     \item[3] Iterative preference learning from human feedback: Bridging theory and practice for rlhf under kl-constraint.
%     \item[4] Insights into alignment: Evaluating dpo and its variants across multiple tasks
%     \item[5] Is DPO superior to PPO for LLM alignment? A comprehensive study
% \end{itemize}

\section{Conclusion}
This paper surveys the most up-to-date state of knowledge on reinforcement learning enhanced LLMs, attempting to consolidate and analyze the rapidly growing research in this field.
We make a systematic review of the literature, including the basics of RL, popular RL-enhanced LLMs, studies on two reward model-based RL techniques—RLHF and RLAIF—and works focused on bypassing the reward model to directly align LLM outputs with human expectations through DPO.
We hope this work will help researchers understand the current challenges and advancements, and motivate further endeavors to address the deficiencies of current RL-enhanced LLMs.

\bibliography{anthology,custom}

\appendix

\end{document}